\documentclass{article}

\usepackage{arxiv}

\usepackage[utf8]{inputenc} 
\usepackage[T1]{fontenc}    
\usepackage{hyperref}       
\usepackage{url}            
\usepackage{booktabs}       
\usepackage{amsfonts}       
\usepackage{nicefrac}       
\usepackage{microtype}      
\usepackage{lipsum}		
\usepackage{graphicx}
\usepackage{natbib}
\usepackage{doi}
\usepackage{algorithm,algcompatible,amsmath}
\usepackage[caption=false,font=normalsize,labelfont=sf,textfont=sf]{subfig}
\algnewcommand\INPUT{\item[\textbf{Input:}]}%
\algnewcommand\OUTPUT{\item[\textbf{Output:}]}%

\title{A Machine Learning Generative Method for Automating Antenna Design and Optimization}

\author{ \hspace{1mm}{Yang~Zhong} \\
	Department of ECE, Duke University\\
	Durham, NC 27708, USA \\
	\texttt{yang.zhong@duke.edu} \\
	\And
	\hspace{1mm}{Peter~Renner} \\
	Reality Laboratory, Meta\\
	Sunnyvale, CA 94087, USA \\
	\texttt{peterrenner@fb.com} \\
	\And
	\hspace{1mm}{Weiping~Dou}	\thanks{This project is sponsored by Facebook (now Meta) Internship Program.} \\
	Reality Laboratory, Meta\\
	Sunnyvale, CA 94087, USA \\
	\texttt{wdou@fb.com} \\
	\And
	\hspace{1mm}{Geng~Ye} \\
	Reality Laboratory, Meta\\
	Sunnyvale, CA 94087, USA \\
	\texttt{gye@fb.com} \\
	\And
	\hspace{1mm}{Jiang~Zhu} \\
	Reality Laboratory, Meta\\
	Sunnyvale, CA 94087, USA \\
	\And
    \hspace{1mm}{Qing~Huo~Liu} \\
	Department of ECE, Duke University\\
	Durham, NC 27708, USA \\
	\texttt{qhliu@duke.edu} \\
}

\date{}

\renewcommand{\shorttitle}{\textit{arXiv} Template}

\hypersetup{
pdftitle={A template for the arxiv style},
pdfsubject={q-bio.NC, q-bio.QM},
pdfauthor={David S.~Hippocampus, Elias D.~Striatum},
pdfkeywords={First keyword, Second keyword, More},
}

\begin{document}
\maketitle

\begin{abstract}
	To facilitate the antenna design with the aid of computer, one of the practices in consumer electronic industry is to model and optimize antenna performances with a simplified antenna geometric scheme. Traditional antenna modeling requires profound prior knowledge of electromagnetics in order to achieve a good design which satisfies the performance specifications from both antenna and product designs. The ease of handling multi-dimensional optimization problems and the less dependence on the engineers' knowledge and experience are the key to achieve the popularity of simulation-driven antenna design and optimization for the industry. In this paper, we introduce a flexible geometric scheme with the concept of mesh network that can form any arbitrary shape by connecting different nodes. For such problems with high dimensional parameters, we propose a machine learning based generative method to assist the searching of optimal solutions. It consists of discriminators and generators. The discriminators are used to predict the performance of geometric models, and the generators to create new candidates that will pass the discriminators. Moreover, an evolutionary criterion approach is proposed for further improving the efficiency of our method. Finally, not only optimal solutions can be found, but also the well trained generators can be used to automate future antenna design and optimization. For a dual resonance antenna design with wide bandwidth, our proposed method is in par with Trust Region Framework and much better than the other mature machine learning algorithms including the widely used Genetic Algorithm and Particle Swarm Optimization. When there is no wide bandwidth requirement, it is better than Trust Region Framework.
\end{abstract}


\section{Introduction}
\label{SecIntro}

Antenna design is considered as one of the most significant hardware engineering challenges in consumer electronic design, not only because of the demand for more antennas and more frequency bands to cover in modern communication system for enabling new use cases and features with ever-growing throughput requirements, but also because of the sophisticated interactions and performance trade-offs between antennas and adjacent electronic components, i.e., display, camera, speaker/mic, sensors, etc, all competing with the shared real-estate constrained from the industrial design of the consumer electronic devices.   
Antenna design usually includes the following procedures: 1) selecting antenna type or shape; 2) specifying and initializing tuning parameters; and 3) modifying the geometrical scheme towards design targets either using simulation tools or experimental prototypes \cite{Behdad2004BandwidthEA,Wong2008WidebandSB,Qing2010ABU,Pazin2011InvertedFLA,Tang2016PlanarUA,Su2019CompactTS}. If no optimal scheme is found in the desired space, the first two procedures will have to be revisited. Therefore, we can simplify the computer aided antenna design process into two procedures: modeling and optimizing. Modeling extracts design parameters from the antenna geometric scheme. Optimizing uses algorithms to find a best set of parameter values whose simulation results will meet the design targets. A right number of design parameters is a trade-off between complexity of modeling and optimizing. For example if we only model length and width for the planar inverted-F antenna, we might not be able to meet all the antenna targets because we lose the flexibility of tuning the other design parameters such as height, material properties and etc.  To come up a more delicate geometric model with a certain degrees of freedom is challenging. 

With the increasing computational power and booming optimizing algorithms, we believe modeling scheme could be delicately simplified. We desire an antenna modeling scheme which requires minimum antenna domain knowledge while having high confidence in the existence of solutions. A novel antenna modeling scheme is therefore proposed in this work. Its geometric model is parameterized by the coordinates and radii of a group of nodes. It also can be flexibly reconfigured by adding free nodes and prior simulation results are still valid.

For optimizing antenna geometric schemes, there are a few popular algorithms including the Trust Region Framework \cite{Alexandrov1997ATF}, Nelder Mead Simplex Algorithm \cite{Dennis1985OptimizationOM}, Covariance Matrix Adaption (CMA) Evolutionary Strategy \cite{Hansen2016TheCE}, Genetic Algorithm \cite{Goldberg1988GeneticAI} and Particle Swarm Optimization \cite{Kennedy1995ParticleSO}. Each algorithm has its own characteristics in terms of global/local convergence, dependency on the initial model, and ability for single/multiple objectives. The general procedures of these algorithms can be summarized: 1) simulating an initial geometric model or a set of initial models; 2) proposing new candidate(s) based on previous simulation results; 3) simulating the new candidates and looping with step 2) until finish. Their major difference is how to acquire new candidates. 

Meanwhile, a few machine learning algorithms are emerging in antenna area \cite{Burrascano1999ARO,Zhang2003ArtificialNN,ZHU2007TAP_Special_issue, Pastorino2005ASA,Xiao2018MultiparameterMW,Cui2020AME,9395365,Sharma2020MachineLT,Wu2020MultistageCM,Nan2021DesignOU,Davoli2021MachineLA,Zhou2021ATP,Naseri2021AGM,Koziel2021AccurateMO,Abdullah2022SupervisedLearningBasedDO}. These algorithms can be summarized into two categories: 1) training surrogate models to replace the electromagnetic simulator for acceleration; 2) training inverse design models to propose new candidates. We call the former discriminator, the latter generator. 

Inspired by the generative adversarial networks (GANs) \cite{Goodfellow2014GenerativeAN} which have been widely used in image synthesis, we are proposing a machine learning generative method to optimize the connecting nodes modeling scheme that we are proposing in this paper for antenna modeling. This method integrates the machine learning algorithms in both categories we summarized earlier, and thus it has virtues from both categories: discriminator accelerates the evaluations, and generator improves the quality of new candidates. In contrast to the adversarial relationship in the GAN framework, the generator's performance depends on the accuracy of discriminator in antenna design. We will use the simulated antenna dataset to train a discriminator and expect the discriminator to be as accurate as the results from the simulator. Then we use the discriminator to train a generator and expect its generated candidates to achieve the design targets. 

The proposed antenna generative method not only works well for antenna geometric optimization, but also is able to automate the computer-aided design process when the proposed connecting nodes modeling scheme is being used. The major contributions of this paper include: 1) studying a simplified geometric modeling scheme to reduce engineers' efforts on determining antenna shapes and parameter variables; 2) integrating discriminators and generators into one antenna generative method to facilitate antenna design and increase optimization efficiency; 3) evaluating a potential approach to enable antenna design automation. In the following sections, the new antenna modeling scheme is described in Section \ref{randomGenerator}. Discriminators, generators and the evolutionary approach are discussed in Section \ref{Algorithms}.  Section \ref{Examples} details two examples using the proposed method and shows the comparison results with other existing algorithms. Conclusions are made in Section \ref{secConc}.

\section{Antenna Geometric Modeling Scheme}
\label{randomGenerator}
\begin{figure}[!t]
\centering \includegraphics[width=0.9\columnwidth]{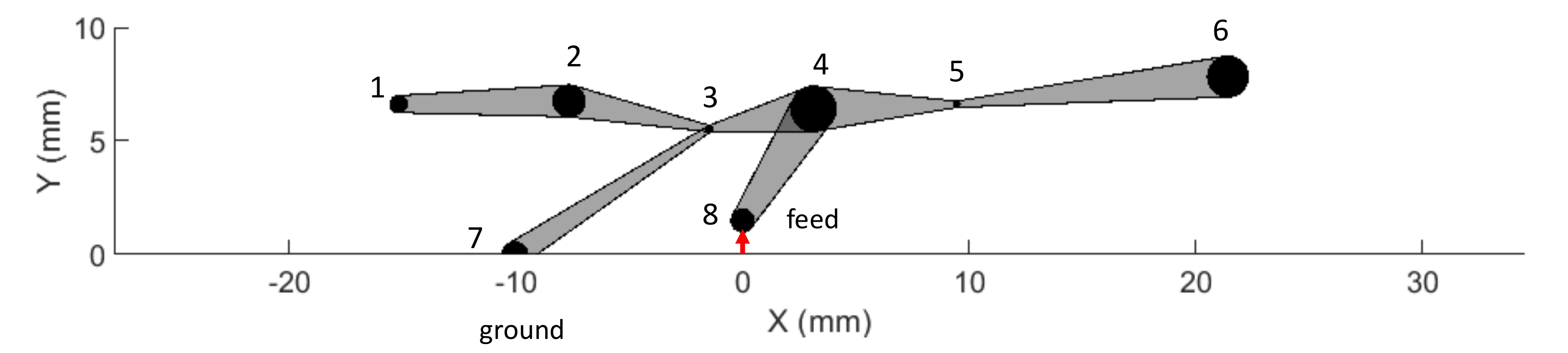}\\
  \caption{Illustration of the connecting nodes antenna modeling scheme. In this case, 20 geometric variables (nodes' coordinates and radii) are adjustable in the predetermined range. Trapezoids connect the pairs of nodes: 1-2,2-3,3-4,4-5,5-6,3-7,4-8. Node 7 connects the ground (Y<0). A feed port is added at Node 8. }\label{connectingNodes}
\end{figure}

As discussed in Section \ref{SecIntro}, we desire a flexible modeling scheme which can easily increase the degrees of freedom of antenna design and does not require much antenna domain knowledge. Fig.~\ref{connectingNodes} illustrates a planar antenna modeling scheme with nodes and trapezoids. The geometric design parameters are the coordinates ($X_i,Y_i$) and radii ($R_i$) of these nodes all on the same plane. Node 7 is the grounding point, and its Y coordinate is fixed ($Y_7 = 0$). Node 8 is the feed point which is also fixed in this example ($X_8 = 0$, $Y_8 = 1.5$ mm, $R_8 = 0.5$ mm). A well defined port where we define $S_{11}$ is added between the ground (Y < 0) and the node 8, the red arrow shown in Fig.~\ref{connectingNodes}. The remaining parameters are variable in a predetermined range, listed in Table I. The current model has 20 degrees of freedom ($X_i,R_i$ where $i\in[1,7]$, $Y_i$ where $i\in[1,6]$) and can be extended by adding additional nodes or enabling unused parameters such as $X_8$, $Y_8$, $R_8$. Once these nodes are determined to be connected, trapezoids will connect them automatically. In this example, the connection map is predefined: nodes 1 to 6 are connected sequentially, nodes 3 and 7 are connected, and nodes 4 and 8 are connected.

\begin{table}[h]
\begin{center}
\caption{Ranges of Design Parameters.} \label{Table1Label}
\begin{tabular}{|c|c|c|}
 \hline
 $X_1 \in$ [-30,-15]\ mm & $X_2\in$ [-15,-5]\ mm & $X_3\in$ [-5,-0]\ mm\\
 \hline
 $X_4\in$ [0,5]\ mm & $X_5\in$ [5,15]\ mm & $X_6\in$ [15,30]\ mm\\
 \hline
 $X_7\in$ [-30,0]\ mm & $Y_{1:6}\in$ [3,10]\ mm & $R_{1:7}\in$ [0.1,1.5] \ mm\\
 \hline
\end{tabular}
\end{center}
\end{table}

\begin{figure}[!t]
\centering \includegraphics[width=1 \columnwidth]{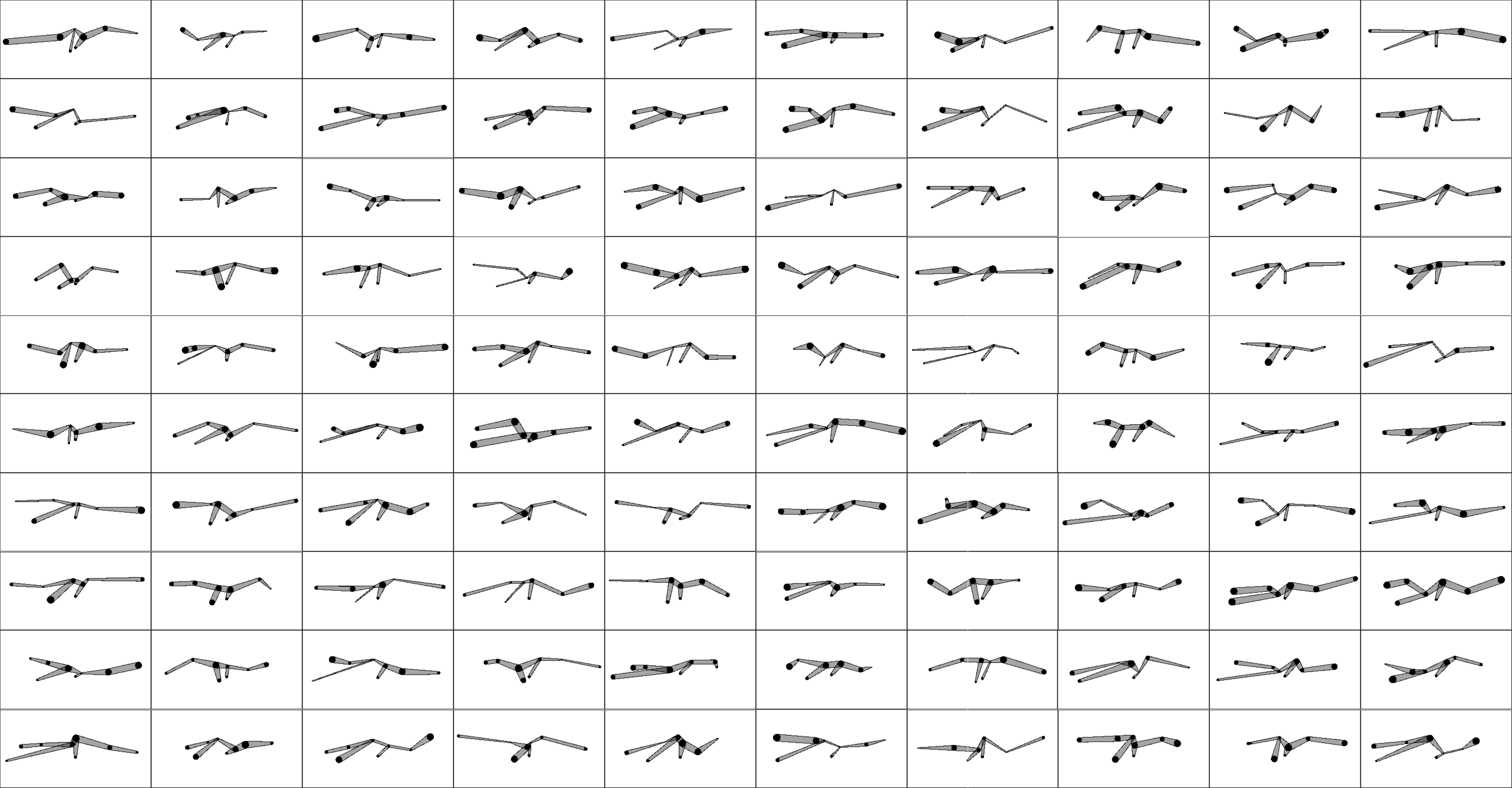}\\
  \caption{Overview of the initial 100 antennas from the simple random antenna generator. The values of 20 geometrical variables are drawn from the uniform distribution. A few cases that form crossing connection lines or shorting feed have been removed by a geometric checker.}\label{antennas_initial100}
\end{figure}

\begin{figure}[!t]
\centering \includegraphics[width=0.9 \columnwidth]{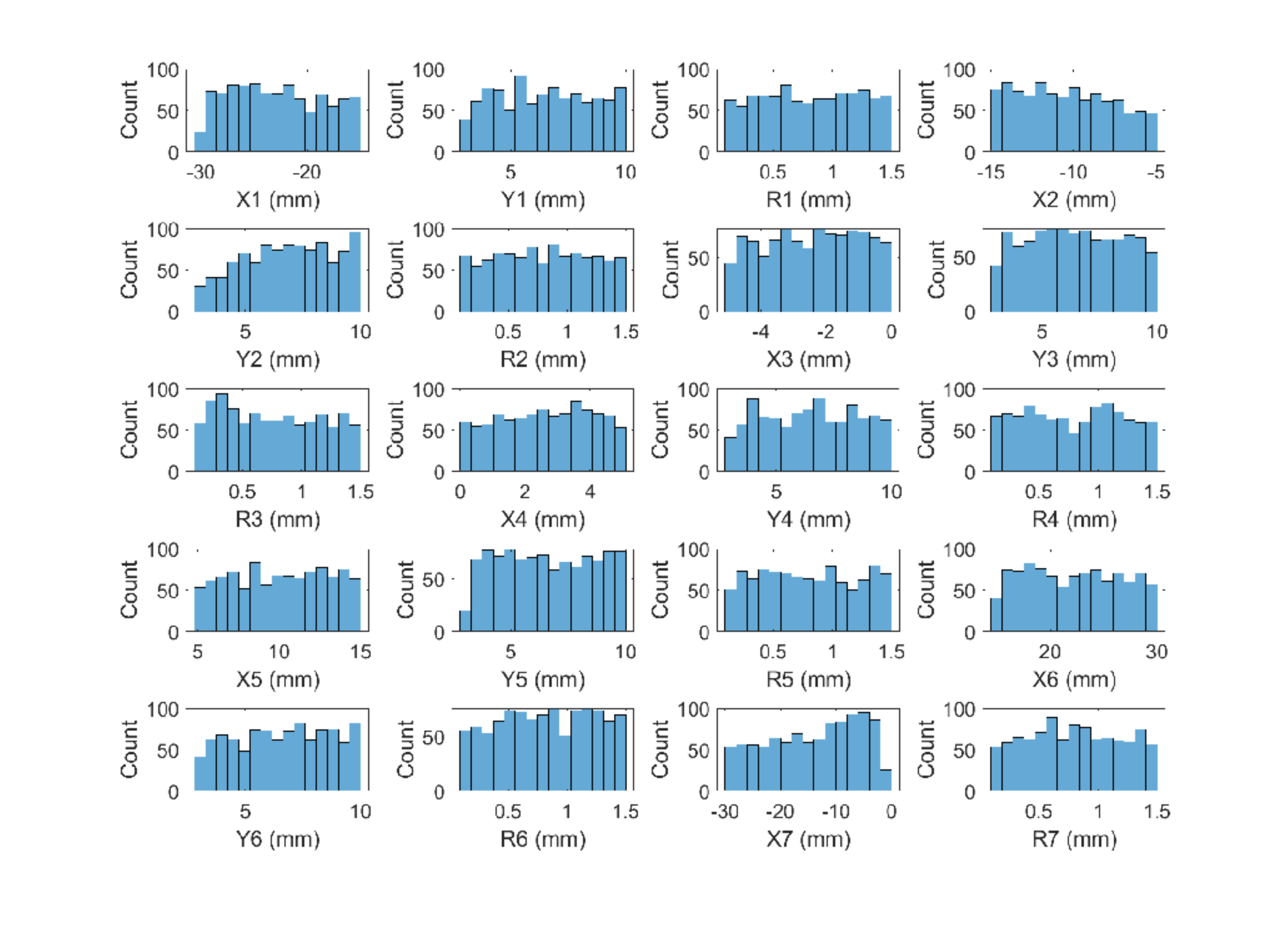}\\
  \caption{The histograms of geometric parameters based on the statistics of the 1000 random antennas that passed the geometric checker. The distributions are almost uniform in their specified ranges.}\label{hist_random1000}
\end{figure}

With such a modeling scheme, we abstracted 20 parameters for characterizing the planar antenna geometry. Each parameter can be independently assigned to a random values drawn from a uniform distribution. A simple random antenna generator is created for generating design candidates though they may or may not achieve the design targets. Additionally, we moved singular cases like crossing connection lines or shorting feed by using a geometric checker. Fig.~\ref{antennas_initial100} shows the 100 random samples generated by the simple random antenna generator that have passed the geometric checker. The histograms of the 20 design parameters are illustrated in Fig.~\ref{hist_random1000}. Strictly speaking, the distributions in Fig.~\ref{hist_random1000} are not shown perfect uniform because of the geometric checker.

\section{Algorithms}
\label{Algorithms}

After these above initial antennas are generated and simulated, we will use the following algorithms to achieve the final design target. These include a) antenna performance discriminator, b) potential valid design generator, and c) evolutionary criterion approach.

\subsection{Antenna Performance Discriminator}
\label{SubC1}

For each antenna geometric model, its performance can be estimated by a full wave electromagnetic simulator. After a number of simulation, we obtain a dataset which comprises pairs of antenna geometric parameters and simulated performance metrics. Then we use this dataset to train machine learning models, geometric parameters as input and predicted performance as output. A limited number of dataset might be challenging for capturing the complex mapping accurately from the geometric parameters to the performance metrics ($x \mapsto p$), especially for those higher dimensional spaces with wider variable ranges. Therefore we replace the quantitative metrics with a qualitative term to describe antenna performance, and then the mapping turns out to be much simpler. In this paper, the qualitative term refers to a binary term of valid or invalid, and quantitative term refers to exact $S_{11}$ values.

A binary classifier which maps geometric parameters to a label of valid or invalid ($x \mapsto y$) is a good candidate. The following two equations describe how to convert quantitative values into qualitative ones for the training purpose. 
\begin{eqnarray} \label{criterion1}
y=\left\{
\begin{aligned}
1 \ (\text{valid}), & & \text{if} & & \forall \ p_i \leq c_{i}\\
0 \ (\text{invalid}), & & & & \text{otherwise}  \\
\end{aligned}
\right.
\end{eqnarray}
Equation \eqref{criterion1} uses independent pass/fail criterion for each quantitative metric. The geometric model will be labeled valid only when all the metrics pass their own criterion (assuming $p_i \leq c_{i}$ is desired; $p_i$ is the $i^{th}$ performance; $c_{i}$ is the criterion for the $i^{th}$ performance). 
\begin{eqnarray} \label{criterion2}
y=\left\{
\begin{aligned}
1 \ (\text{valid}), & & \text{if} & & \sum_{i=1}^{m} w_i p_i \leq c\\
0 \ (\text{invalid}), & & & & \text{otherwise}\\
\end{aligned}
\right.
\end{eqnarray}
In \eqref{criterion2}, the geometric model will be labeled valid only when the weighted ($w_i$) sum of quantitative metrics passes a criterion. The criterion $c_i$ in \eqref{criterion1} and $c$ in \eqref{criterion2} can be either constant values or percentiles of the training data, for example, the median values of the quantitative performance metrics such as $|S_{11}|$. 

After abstracting a binary label from the simulated performance, we reformat the dataset into pairs of geometric parameters and the binary label. Multiple techniques can be applied for characterizing the new mapping. For example, training a neural network with geometric parameters as input and the probability of valid as output by using gradient descent with binary cross entropy loss. We can also use support vectors to estimate the boundary in the space of geometric parameters, and thus the support vector classifier (SVC) could train faster. Other binary classification algorithms such as decision trees, k-nearest neighbors can also be used to train the antenna discriminator. No matter which technique is used, overfitting needs to be avoided.

The acceptance rate is low when we use a simple random antenna generator in Section \ref{randomGenerator} to create candidates for this discriminator to filter out invalid models. Next, we will discuss how to create a more efficient generator.

\subsection{Potential Valid Design Generator}
\label{SubC2}
The discriminator $\bf{D}$ draws a rough outline of potential valid design space based on the training dataset. The geometric model $x$ predicted to be valid ($ {\bf{D}} (x)=1$) does not guarantee that it is actually valid ($p \leq c$), while its possibility is higher than a random guess. Therefore, we propose to use the discriminator to train a generator which outputs to potential valid geometric models. 

A generator $\bf{G}$ mapping a random noise vector $z$ to the potential valid geometric model $\hat{x}$ ($z \mapsto \hat{x}$) is used:
\begin{eqnarray} \label{generator}
\hat{x} = {\bf{G}}(z), \  s.t. \  {\bf{D}}(\hat{x})=1.
\end{eqnarray}
The generator $\bf{G}$ is going to solve a regression problem. Its output is a vector of geometric parameters which are continuous in the design ranges. We can train a feed-forward neural network by descending the gradient:
\begin{eqnarray} \label{gradient}
\nabla_{\theta_g}(\frac{1}{m} \sum^{m}_{i=1} {\rm log}(1-{\tilde{\bf{D}}}({\bf{G}}(z^{(i)}))) ),
\end{eqnarray}
where $\theta_g$ represents the trainable parameters in the generator's neural network. The discriminator in \eqref{gradient} is denoted by $\tilde{\bf{D}}$, which has small differences in terms of the output. Here, the output of $\tilde{\bf{D}}$ is a float number in the range of $(0,1)$, indicating the probability of predicting a valid design. The training minibatch size is $m$. The input of generator is $z^{(i)}$ which is an independent random vector. During the whole training process, the random vector $z^{(i)}$ is always a new sample from a predetermined distribution $p_g(z)$. The prior distribution could be uniform, normal or any other type. The generator is to map a prior distribution of the noise vector $z$ to a fancy distribution of the potential valid geometric parameters.

The training procedures are summarized in the Algorithm 1. Both the discriminator and generator are implemented by the feed-forward neural networks. The uppercase characters such as $X$ and $P$ represent sets with $n$ samples; the lowercase characters such as $x^{(i)}$ and $p^{(i)}$ represent the $i^{th}$ sample's geometric parameters and performance metrics, respectively. From this algorithm, we observe that the quality of the generator depends on the test accuracy of the discriminator. The higher accuracy that the discriminator is able to achieve, the higher percentage of true valid designs that the generator generates. 

\begin{algorithm}
    \caption{Proposed Generative Algorithm}
  \begin{algorithmic}[1]
    \REQUIRE geometric models $X \{x^{(1)},x^{(2)},...,x^{(n)}\}$ and the corresponding performance metrics $P \{p^{(1)},p^{(2)},...,p^{(n)}\}$
    \INPUT the dataset with $X$ and $P$, performance criterion $c$
    \OUTPUT potential valid geometric models $\hat{X}$
    \STATE Split the geometric models into valid and invalid classes by labeling $Y\{y^{(1)},y^{(2)},...,y^{(n)}\}$ where $y^{(i)}=1$ if $p^{(i)} \leq c$ else $y^{(i)}=0$
    \STATE \textbf{Discriminator} ${\tilde{\bf{D}}}$: train a binary classification neural network (input geometric model, output probability of valid design):
    \FOR{number of training iterations}
    \STATE Update discriminator by descending the gradient:
    \begin{center}
    $\nabla_{\theta_d}(\frac{1}{n} \sum^{n}_{i=1} y^{(i)}  {\rm log}({\tilde{\bf{D}}}(x^{(i)})) + (1-y^{(i)})  {\rm log}(1-{\tilde{\bf{D}}}(x^{(i)})) )$
    \end{center}
    \ENDFOR
    \STATE \textbf{Generator} ${\bf{G}}$: train a regression neural network (input random noise $z$, output potential valid geometric model):
    \FOR{number of training iterations}
    \STATE Sample minibatch of $m$ noise samples $\{ z^{(1)}, z^{(2)},...z^{(m)}\}$ from noise prior $p_g(z)$.
    \STATE Update generator by descending the stochastic gradient:
    \begin{center}
    $\nabla_{\theta_g}(\frac{1}{m} \sum^{m}_{i=1} {\rm log}(1-{\tilde{\bf{D}}}({\bf{G}}(z^{(i)}))) )$
    \end{center}
    \ENDFOR
    \STATE Use ${\bf{G}}({Z})$ to output the potential valid designs $\hat{X}$.
  \end{algorithmic}
\end{algorithm}

In subsection \ref{SubC3}, we will discuss two approaches for further improving the test accuracy: 1) optimizing the training hyperparameters such as network structures, learning rate and regularization, and 2) improving the quality of the training dataset. 

\subsection{Evolutionary Criterion Approach}
\label{SubC3}
In order to obtain an effective potential valid generator, we treat our discriminator carefully. For example, if we desire a generator to output to a design with $|S_{11}| < -10$ dB at certain frequencies, the discriminator with a valid criterion $|S_{11}| < -3$ dB will not be appropriate. The reason is that using the criterion $|S_{11}| < -3$ dB,  the generated candidates are only prepared for satisfying $|S_{11}| < -3$ dB. It is possible that many of them may satisfy $|S_{11}| < -5$ dB but not $|S_{11}| < -10$ dB. It is also possible that the performance of samples obtained by the simple random generator may be far away from the design targets. For example, none satisfies $|S_{11}|<-10$ dB and 50\% $|S_{11}|<-3$ dB. For such dataset, the performance criterion cannot be $|S_{11}| < -10$ dB, while $|S_{11}| < -3$ dB is appropriate. 

\begin{figure}[!t] 
\centering \includegraphics[width=0.9\columnwidth]{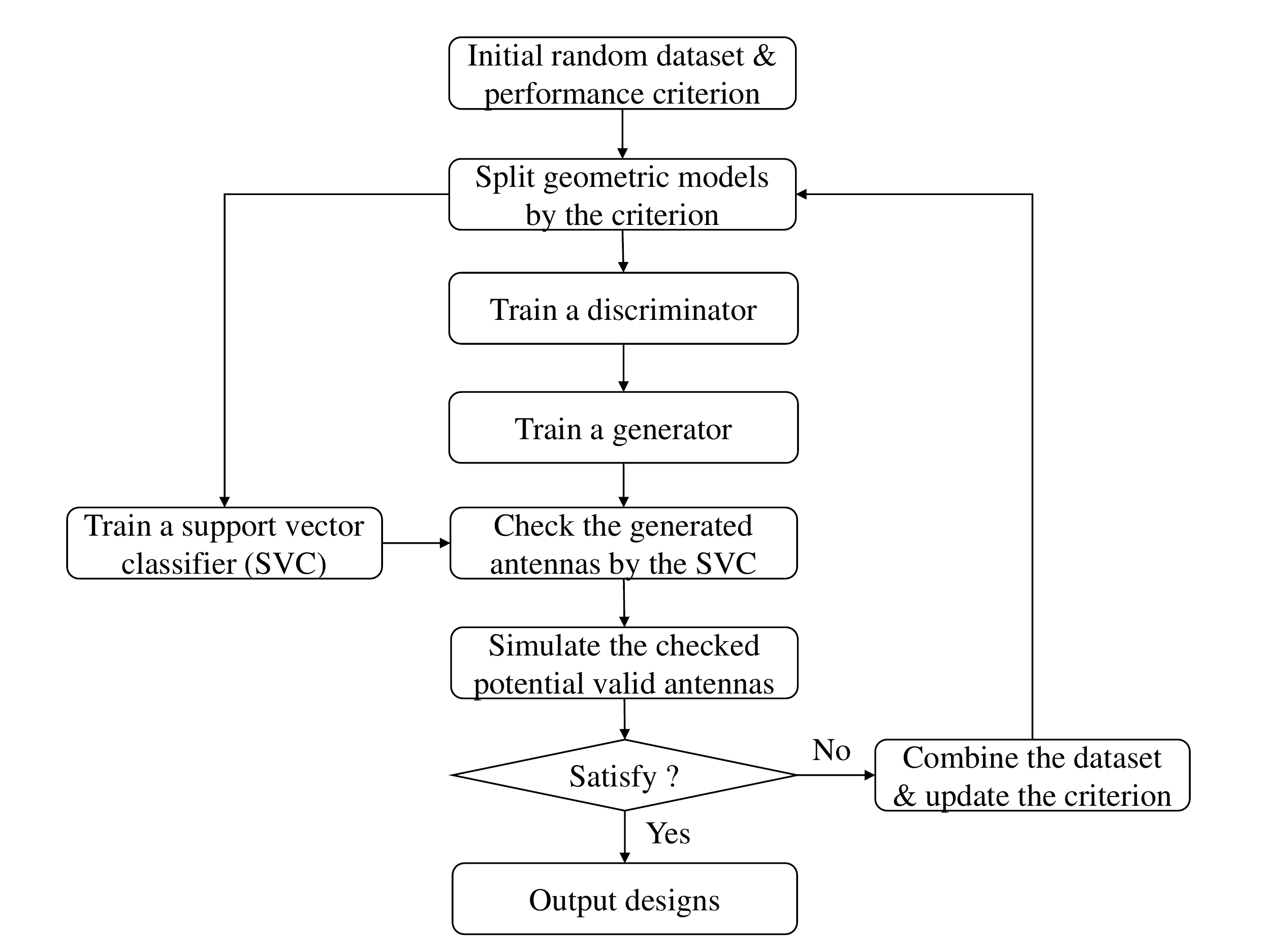}\\
  \caption{The proposed evolutionary strategy with the proposed generative algorithm. A weak criterion such as $|S_{11}|<-3$ dB can be used at the beginning. As the dataset expands by simulating new generated antennas, the criterion should be upgrade towards the design targets such as $|S_{11}|<-10$ dB. The evolution of discriminators and generators is associated with the growing training dataset and stricter criterion. The discriminator and generator are trained and used as described in the Generative Algorithm. The support vector classifier not only double checks the generated antennas, but also suggests a test accuracy that the neural network discriminator should achieve.}\label{algorithm}
\end{figure}

For this, we propose to use an evolutionary approach. Our proposal is to update the discriminator's criterion by iterations. That's to say, using the initial random dataset to train an initial discriminator and an initial generator, the dataset will be expanded after iterations of antenna simulations. Using this strategy, we can keep updating the criterion towards the design goal. The discriminator and generator can also be kept updating by using the new dataset. Such evolution won't stop unless a certain design achieves its target or a desired generator is obtained. For each iteration, the most expensive computational step is simulating the generated candidates and thus it is important to validate them before simulation. We use the most recent dataset to train a supported vector classifier, and then validate the candidates with the classifier. Also, the test accuracy of the support vector classifier is helpful to estimate the test accuracy that the neural network discriminator should achieve. Fig.~\ref{algorithm} illustrates the proposed evolutionary strategy. Note that it is not mandatory to use the same hyperparameters to train the discriminators in each iteration, but they should be tuned for the optimal test accuracy.

There are several advantages of using the proposed evolutionary approach with the proposed generative algorithm: 1) good variety of solutions because the search scope can be changed from global to multiple locals during iterations of the evolution; 2) open to changing design targets because the simulated data especially the initial random dataset is reusable; 3) friendly for parallel acceleration because candidate antennas are independent in each evolution. 

\begin{algorithm}
    \caption{Proposed Evolutionary Criterion Approach}
  \begin{algorithmic}[1]
  \REQUIRE initial random dataset: geometric models $X_0$ and the corresponding performance $P_0$
    \INPUT initial dataset with $X_0$ and $P_0$, initial criterion to split the dataset $c$, the design goal
    \OUTPUT geometric models achieve the goal
    \STATE Use the generative algorithm with initial dataset $X_0$
    \WHILE{not enough designs achieve the goal}
      \STATE Train a support vector classifier (SVC) to filter generated models $\hat{X}$
      \STATE Simulate the SVC accepted models $X_{new}$
      \STATE Combine the dataset $X=\{X,X_{new}\}$
      \STATE Update the criterion $c$
      \STATE Use the generative algorithm with the combined dataset (Adjust hyperparameters when training discriminators)
    \ENDWHILE
    \STATE Output designs achieve the goal
  \end{algorithmic}
\end{algorithm}

\section{Examples}
\label{Examples}

\subsection{Dual Resonance Antenna Design}
\label{Example1}

\begin{figure}[!t]
\centering \includegraphics[width=0.9 \columnwidth]{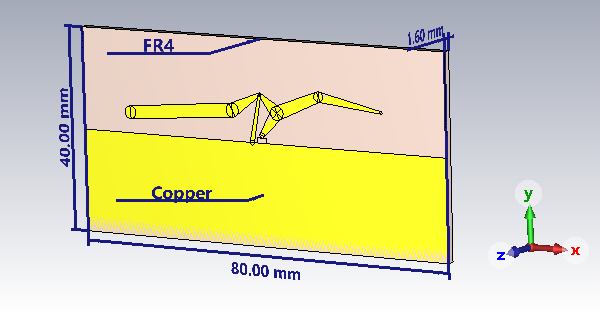}\\
  \caption{A 3D model to be simulated in the CST Studio. The 80 mm $\times$ 40 mm $\times$ 1.6 mm dielectric board is with the material of lossy FR-4. On the top surface of the board, the antenna geometric model and a 80 mm $\times$ 20 mm rectangle are with the material of pure copper. No conductor covers the bottom surface of the board.}\label{3Dmodel}
\end{figure}

The generated antennas will be copper printed on a FR4 board with a size of 80 mm $\times$ 40 mm $\times$ 4 mm and half of the lower surface is copper to serve as the ground plane. Fig.~\ref{3Dmodel} shows one example of the 3D model of these antennas using a commercial EM simulator \cite{CST}. The $|S_{11}|$ parameter of these antenna models will be simulated using CST in the range of 0 to 8 GHz. In this example, a generator will be trained to automatically output to potential valid designs. The design target is specified by: 

\begin{eqnarray} \label{example1}
|S_{11}|_{2.4 \  GHz}+|S_{11}|_{5.9 \  GHz} \leq -20\  dB.
\end{eqnarray}

\begin{figure}
\centering
\subfloat[]{\includegraphics[width=0.3\columnwidth]{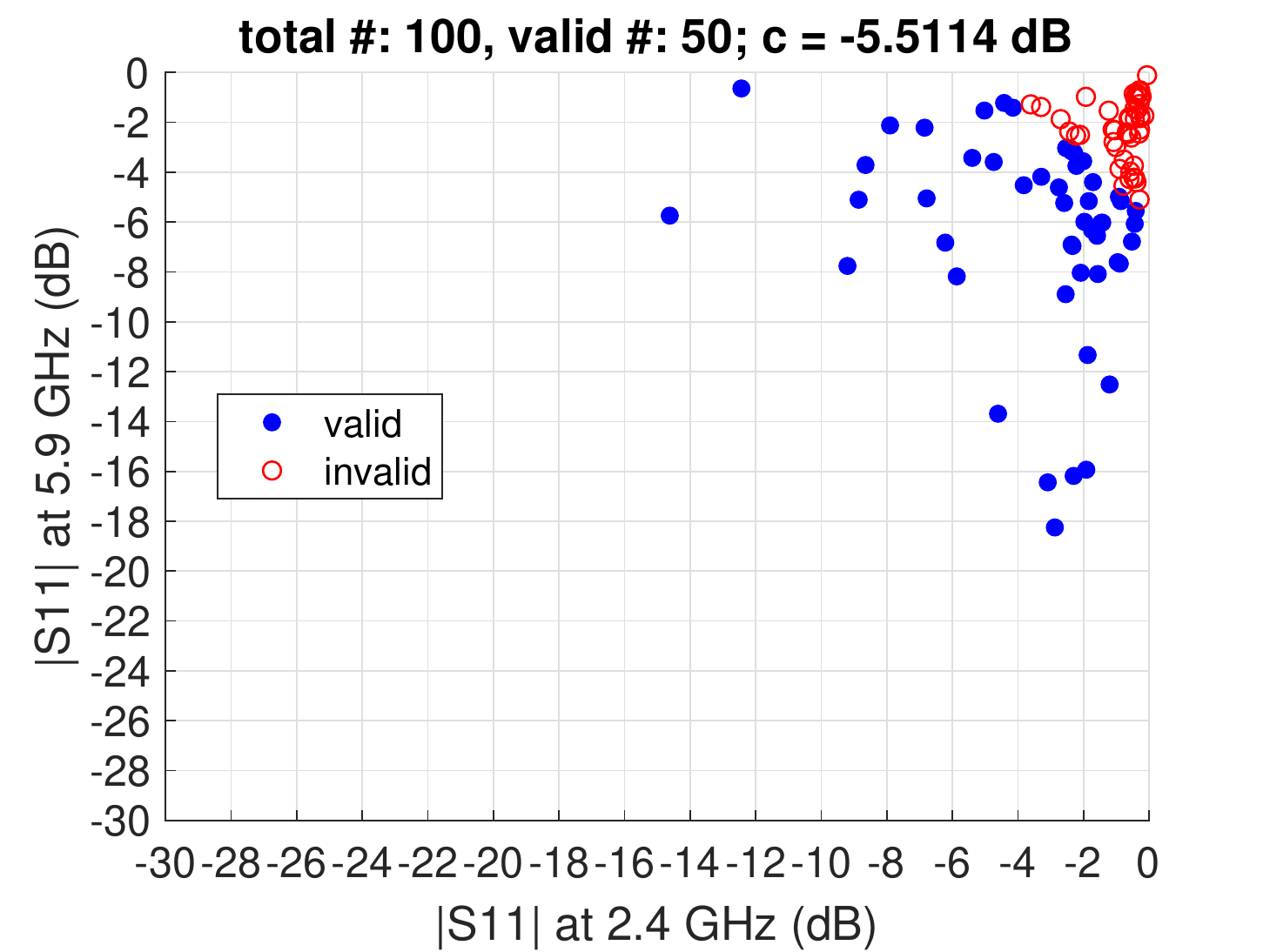}
\label{acc0}}
\subfloat[]{\includegraphics[width=0.3\columnwidth]{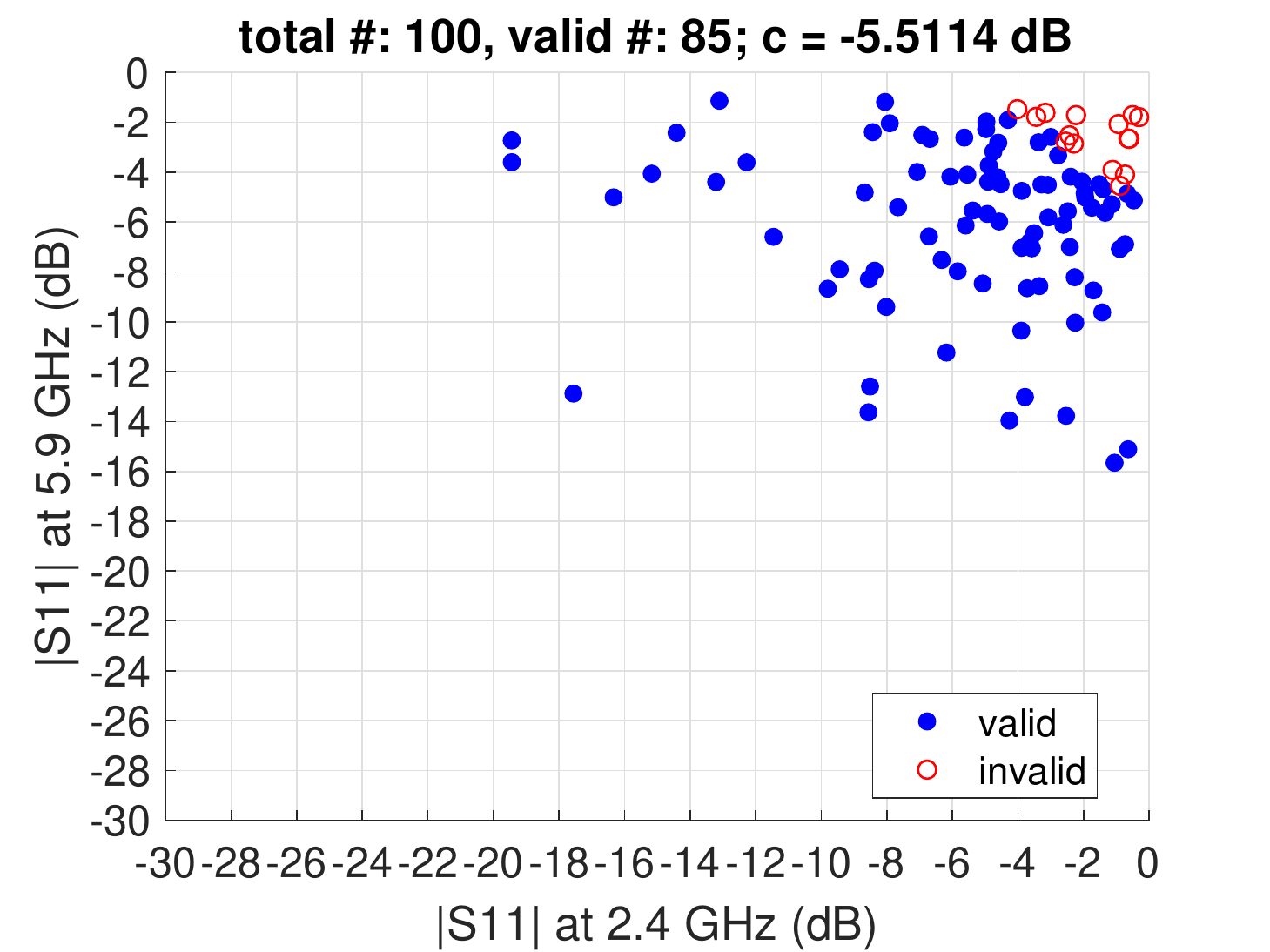}
\label{curr0}}
\subfloat[]{\includegraphics[width=0.3\columnwidth]{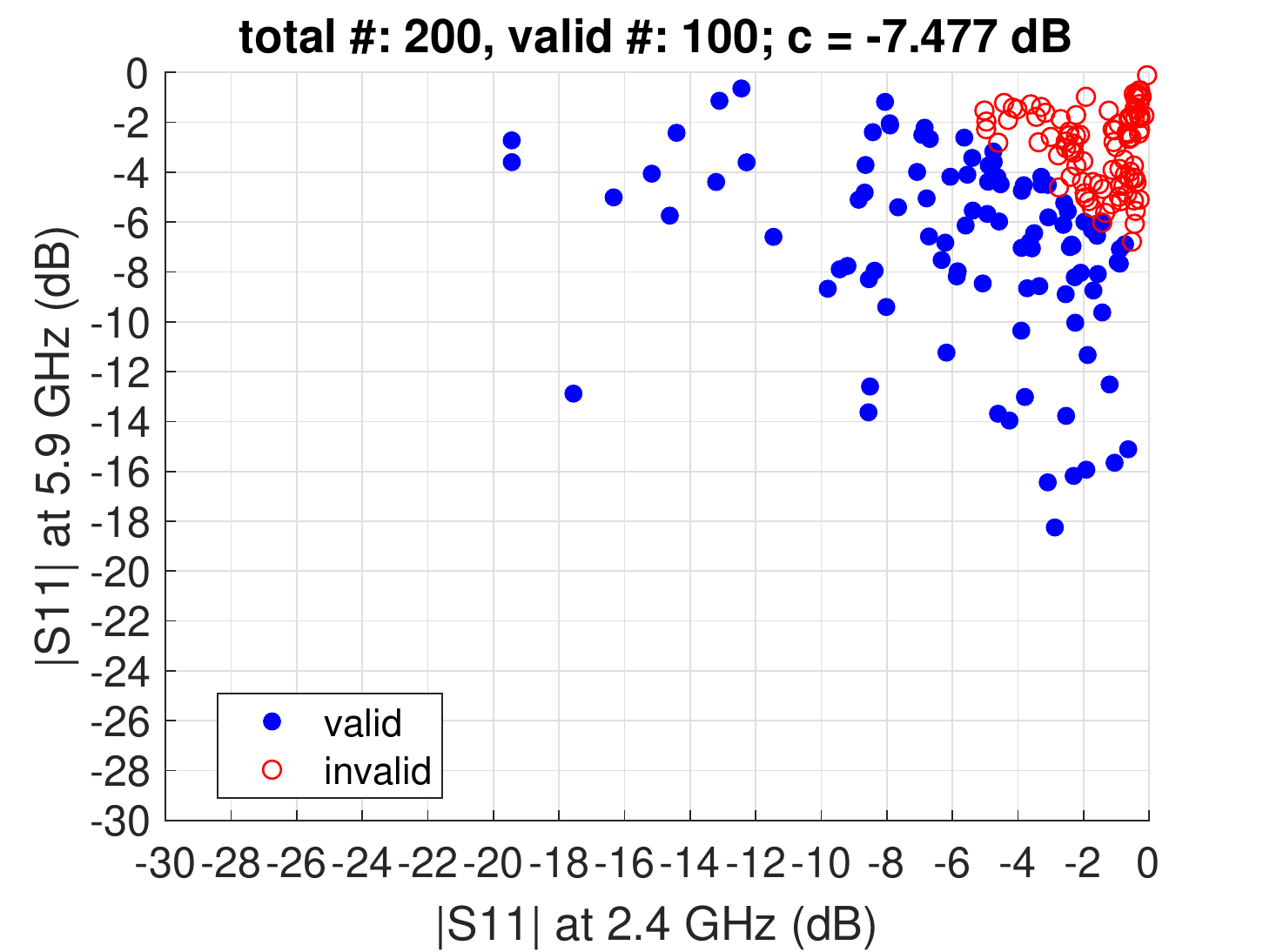}
\label{acc1}}

\subfloat[]{\includegraphics[width=0.3\columnwidth]{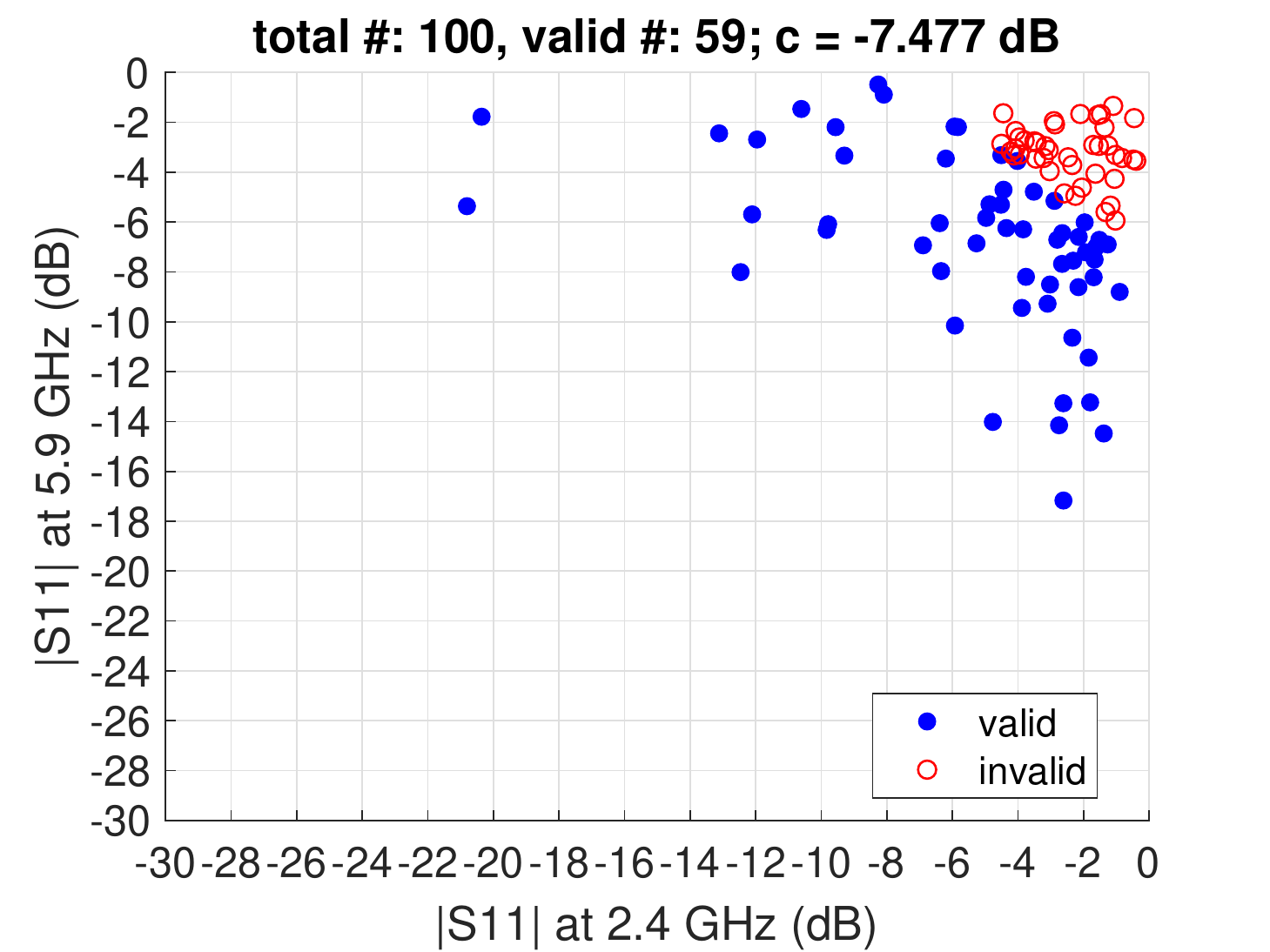}
\label{curr1}}
\subfloat[]{\includegraphics[width=0.3\columnwidth]{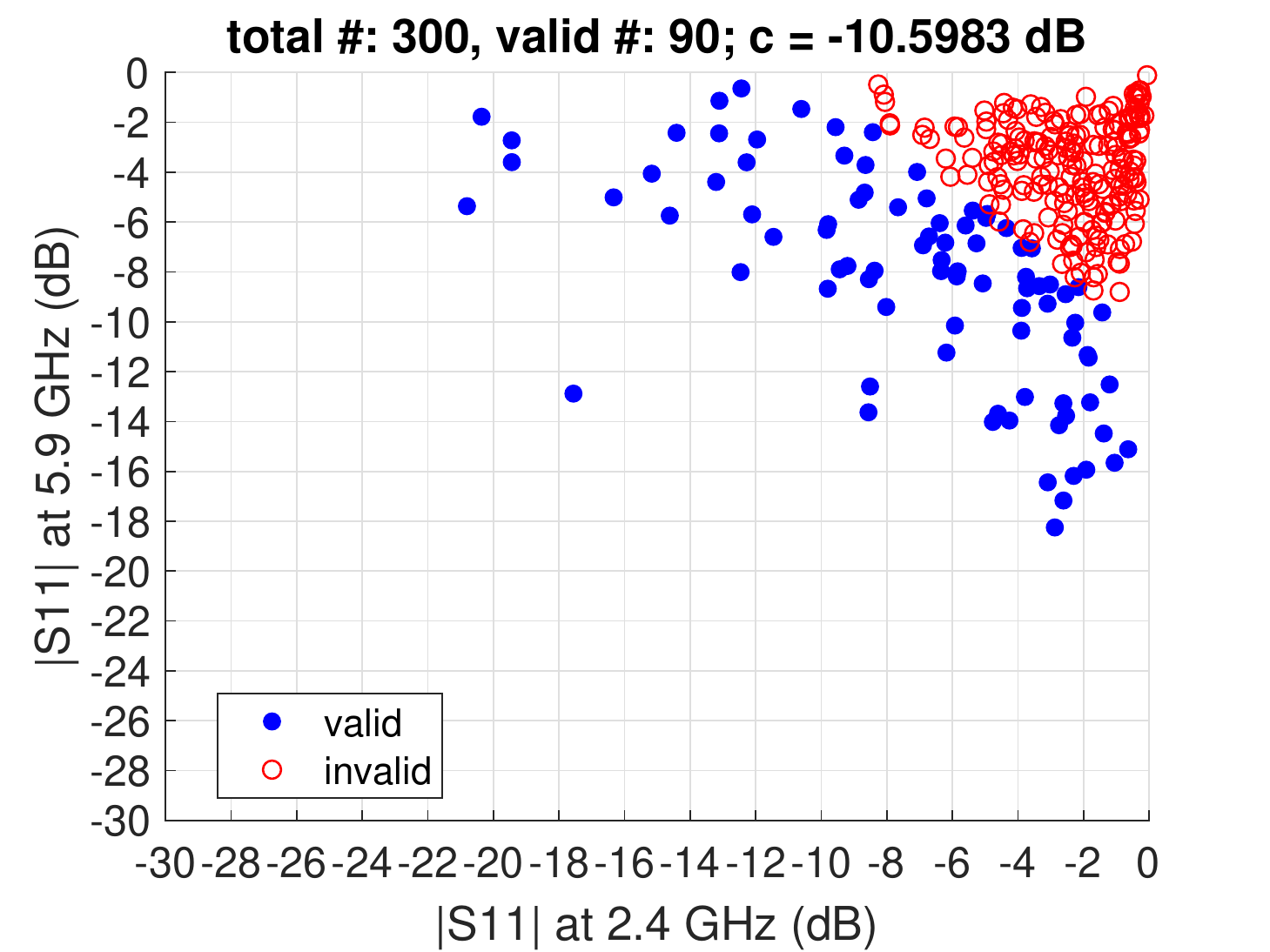}
\label{acc2}}
\subfloat[]{\includegraphics[width=0.3\columnwidth]{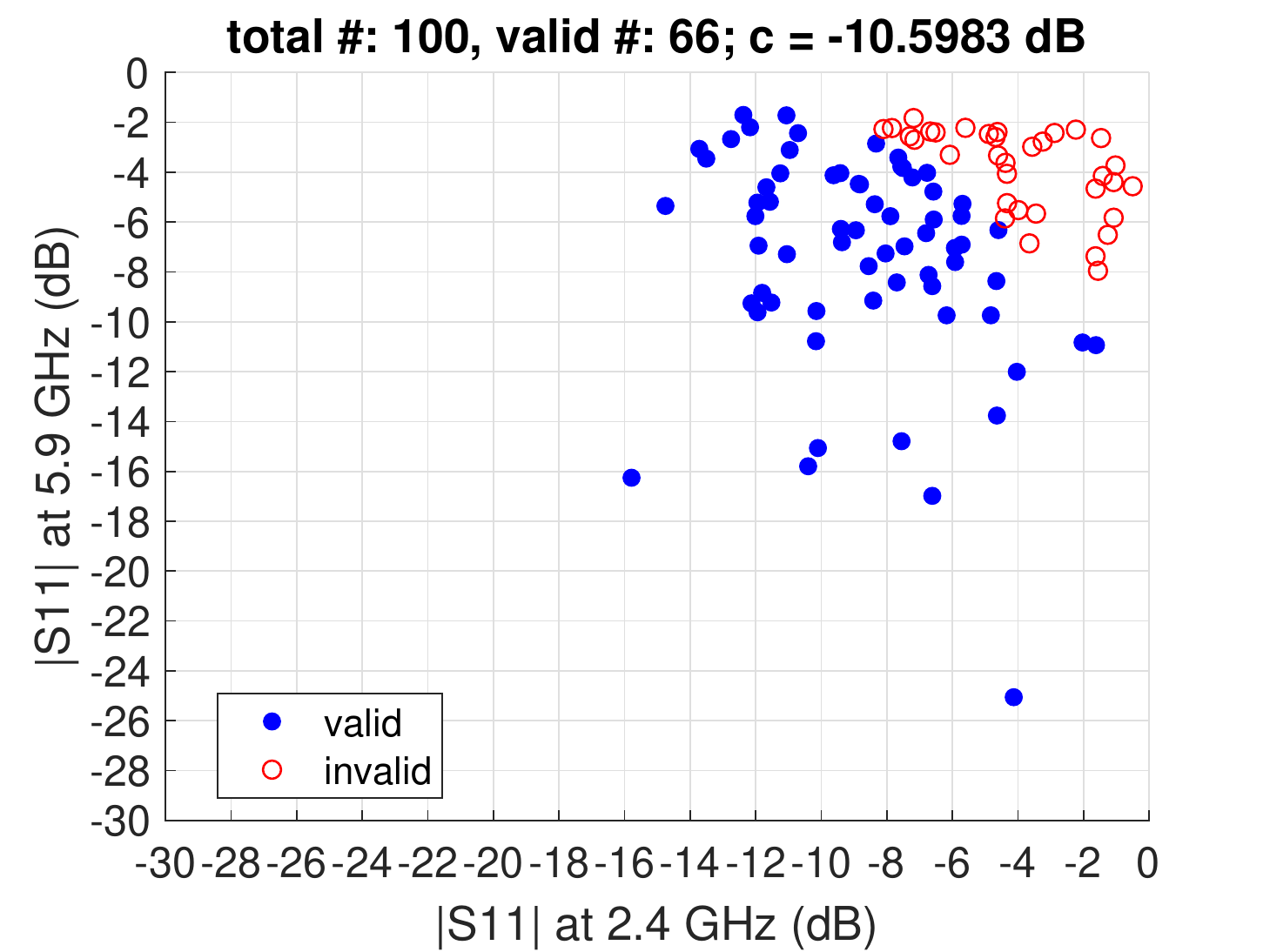}
\label{curr2}}

\subfloat[]{\includegraphics[width=0.3\columnwidth]{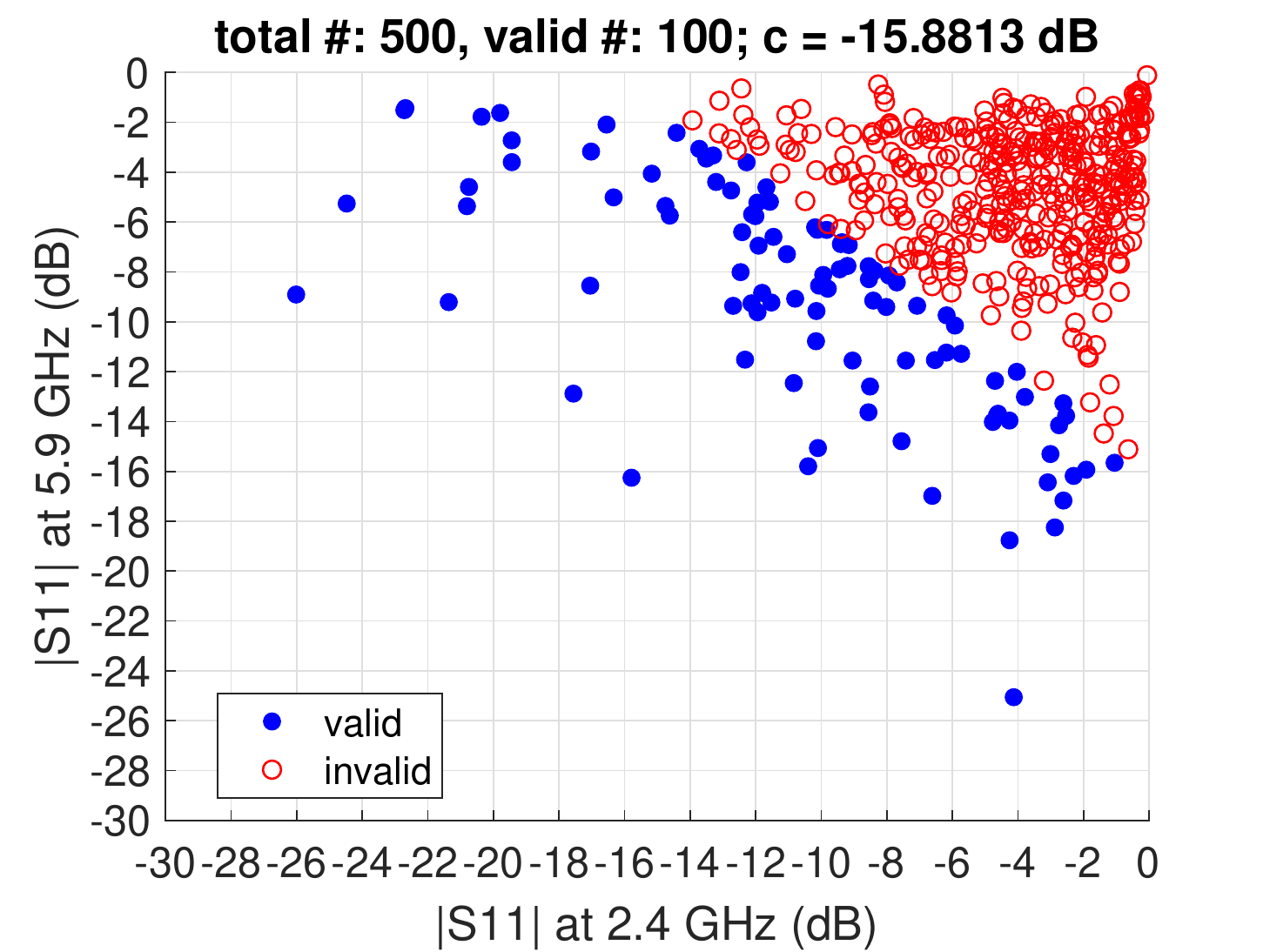}
\label{acc3}}
\subfloat[]{\includegraphics[width=0.3\columnwidth]{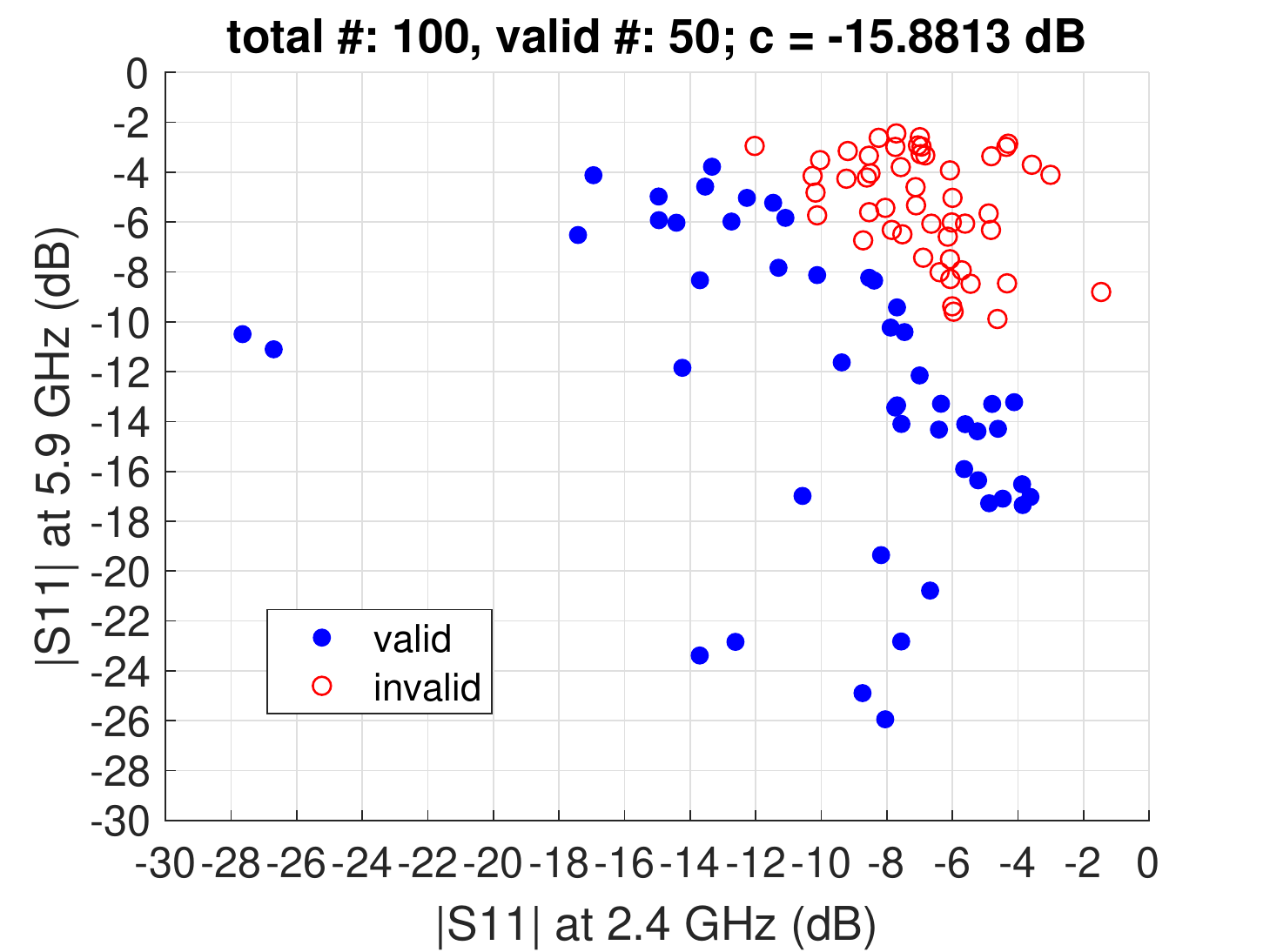}
\label{curr3}}
\subfloat[]{\includegraphics[width=0.3\columnwidth]{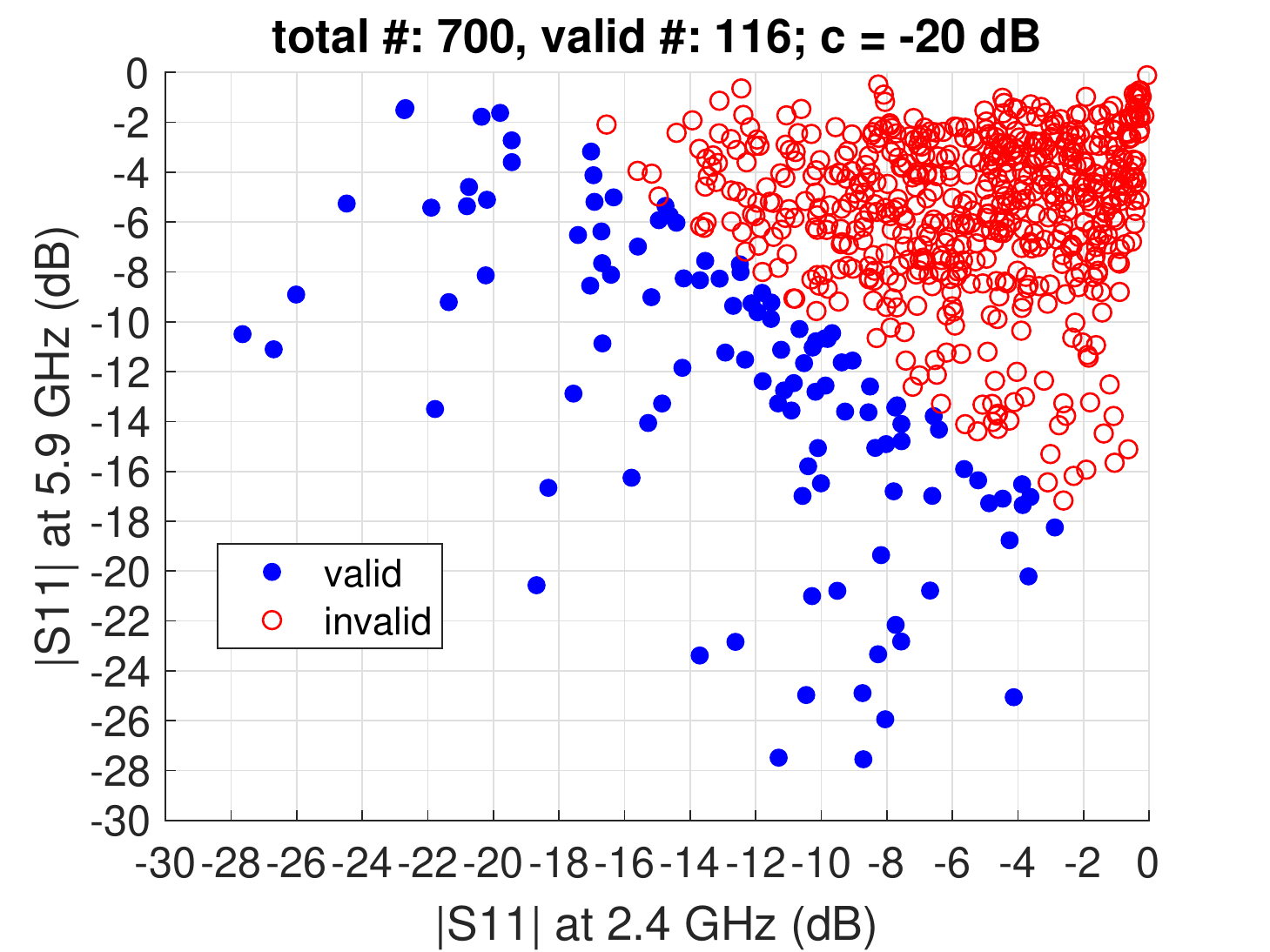}
\label{acc4}}

\subfloat[]{\includegraphics[width=0.3\columnwidth]{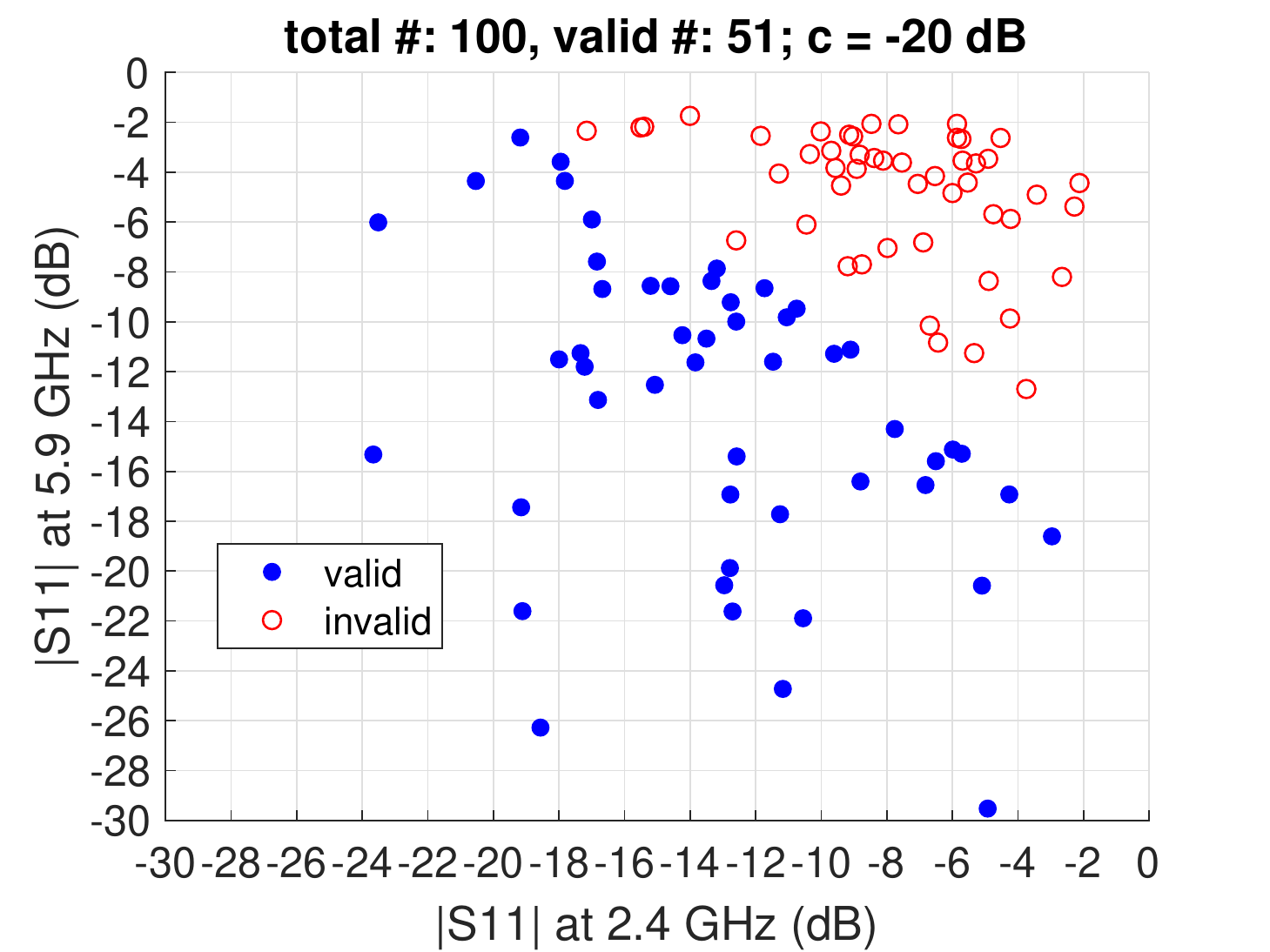}
\label{curr4}}
\caption{Visualization of the dataset in the 2D performance space of the first example. Each dot represents a simulated antenna. (a) The initial dataset in the first evolution. (b) The generated dataset in the first evolution. (c) Combined dataset in the second evolution. (d) The generated dataset in the second evolution. (e) Combined dataset in the third evolution. (f) The generated dataset in the third evolution. (g) Combined dataset in the fifth evolution. (h) The generated dataset in the fifth evolution. (i) Combined dataset in the seventh evolution. (j) The generated dataset in the seventh evolution.}
\label{intermediate}
\end{figure}

First of all, the initial 100 random antennas shown in Fig.~\ref{antennas_initial100} are simulated as baseline. Two of them meet the target of $|S_{11}|<-10$ dB either at 2.4 GHz or 5.9 GHz, but none of them meet both. The initial dataset can be visualized in a 2D performance space as shown in Fig.~\ref{acc0}. The horizontal axis is the $|S_{11}|$ at 2.4 GHz, while the vertical axis is the $|S_{11}|$ at 5.9 GHz. Each dot represents a generated antenna. To split the dataset into valid and invalid classes, the criterion definition \eqref{criterion2} is used. Then the sum of $|S_{11}|$ at 2.4 GHz plus $|S_{11}|$ at 5.9 GHz shall not be larger than a specific constant $c$. 

\begin{figure}[!t]
\centering
\subfloat[]{\includegraphics[width=0.4\columnwidth]{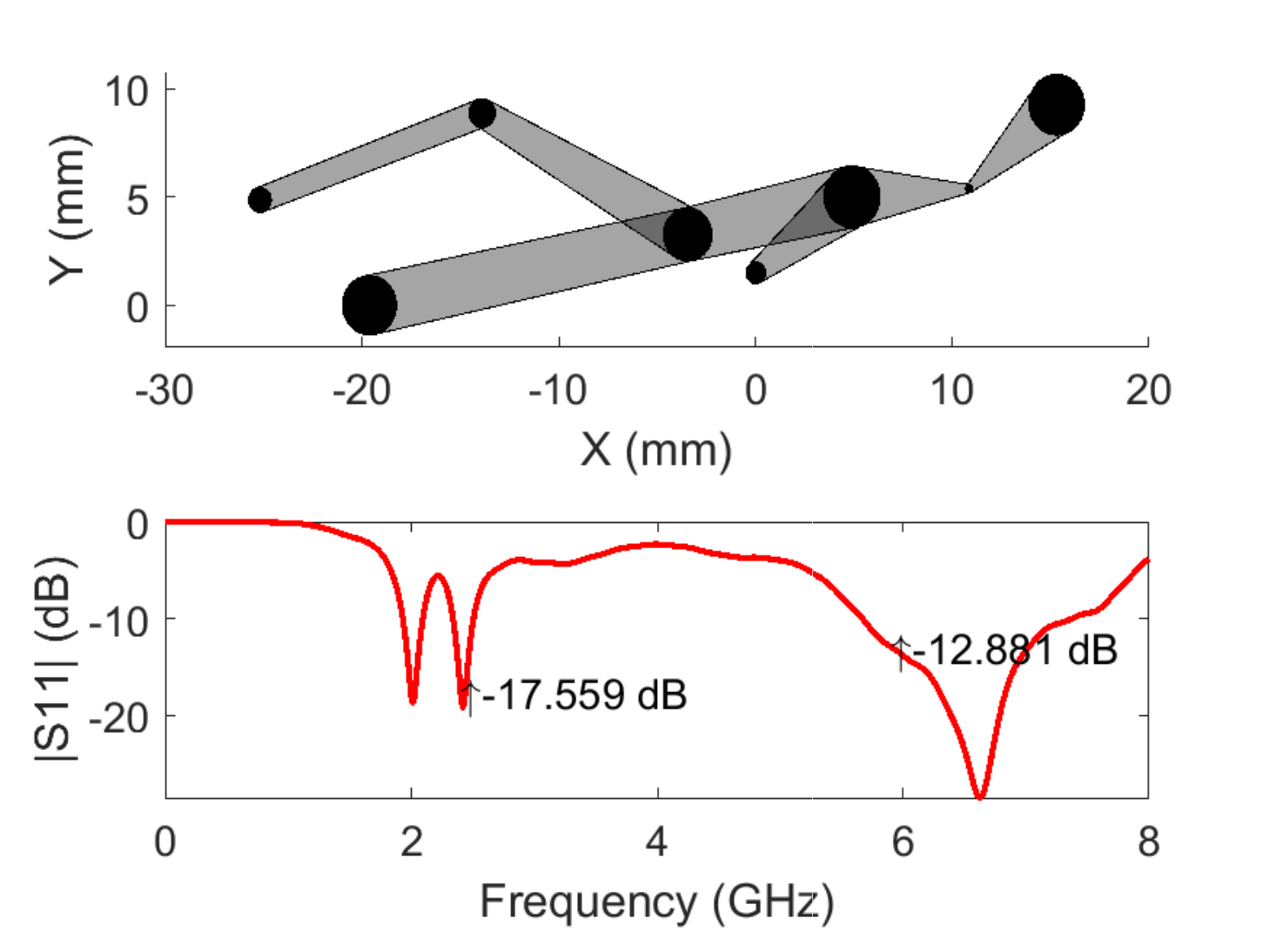}
\label{best1}}
\subfloat[]{\includegraphics[width=0.4\columnwidth]{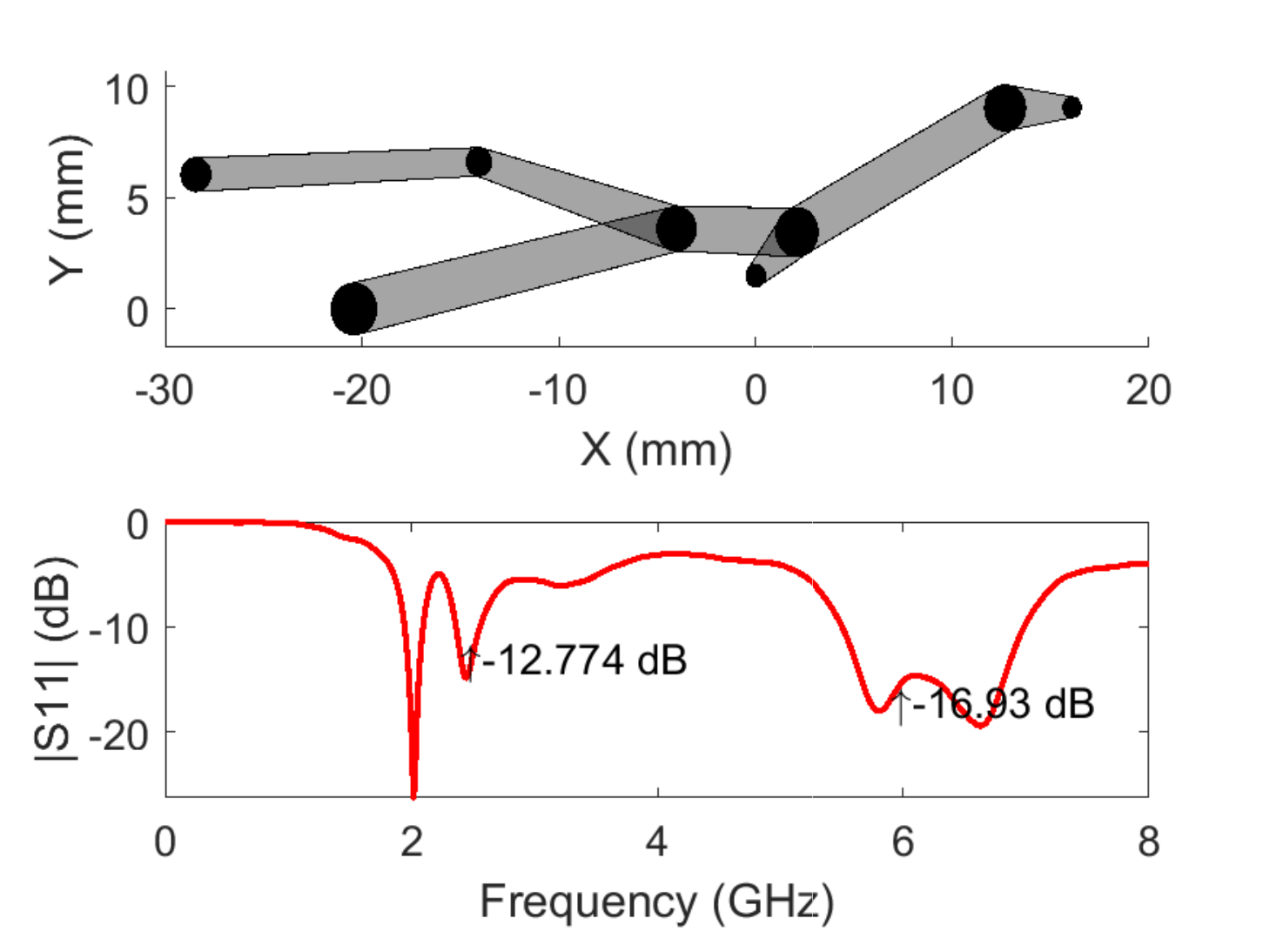}
\label{best2}}

\subfloat[]{\includegraphics[width=0.4\columnwidth]{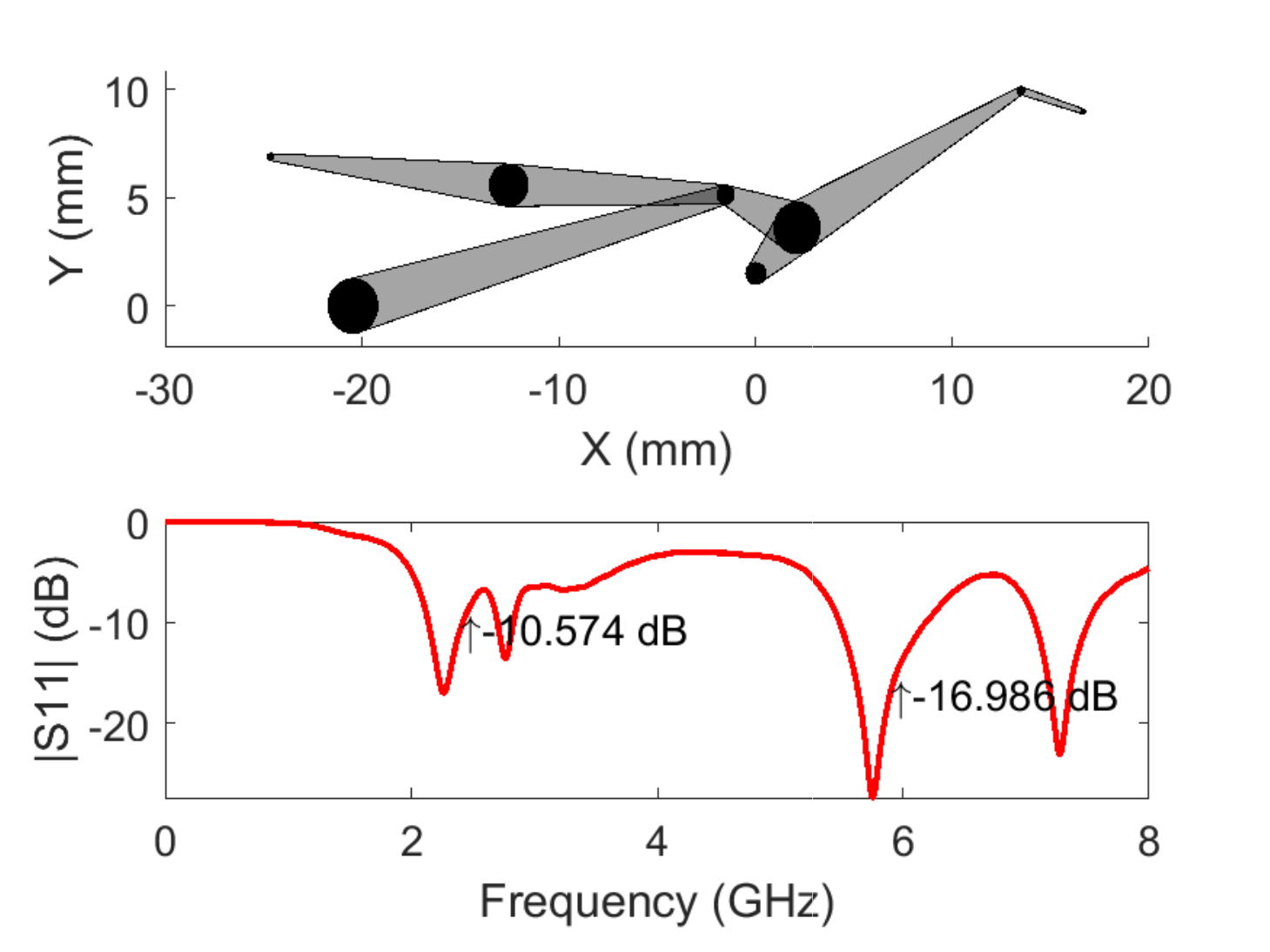}
\label{best3}}
\subfloat[]{\includegraphics[width=0.4\columnwidth]{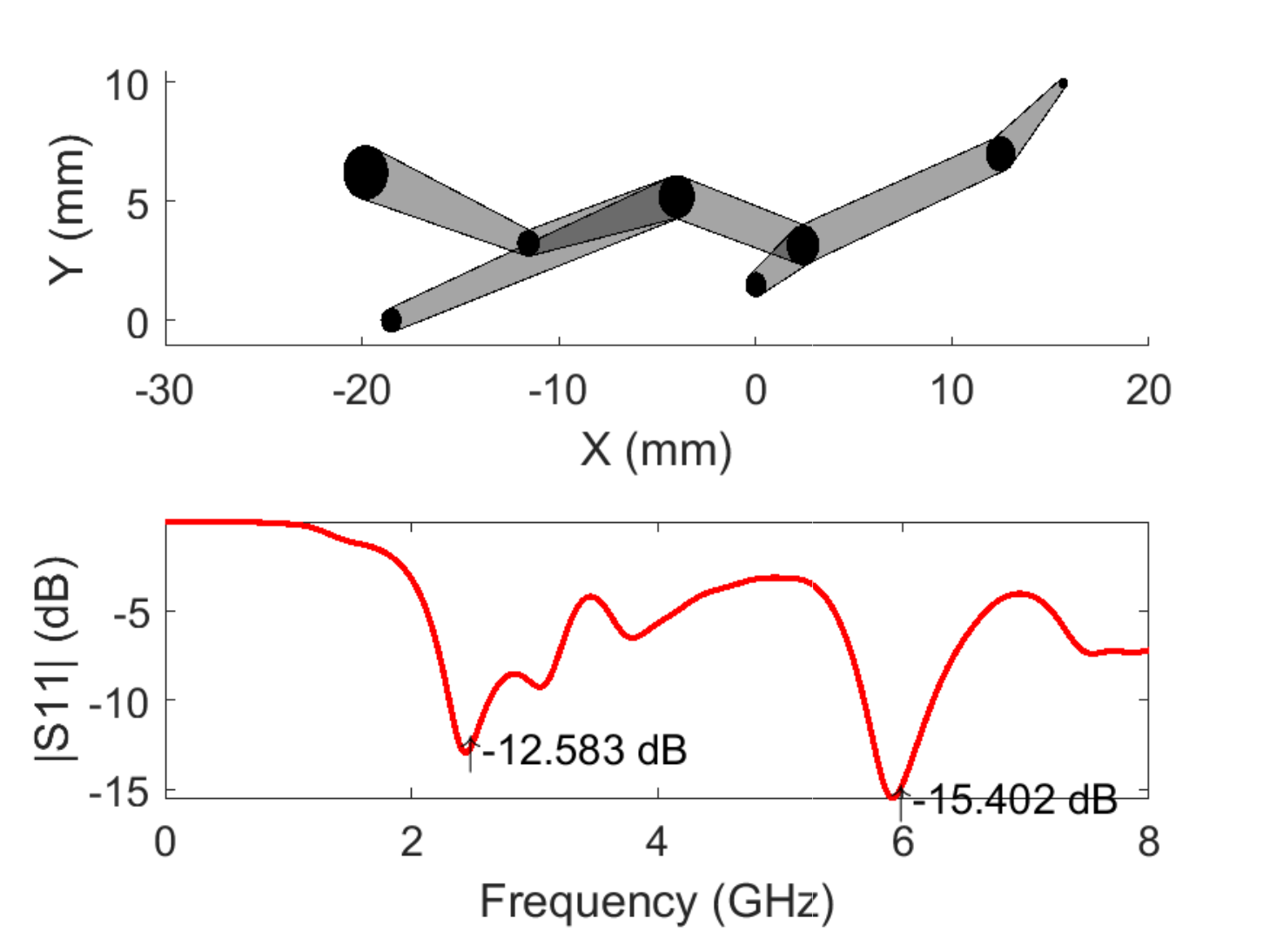}
\label{best4}}
\caption{The antenna geometric model and the corresponding $S_{11}$ curve in the first example. (a) The first design that achieves the dual resonance goal. (b)(c)(d) The other three designs that achieve the dual resonance goal.}
\label{best}
\end{figure}

The constant is determined by the median value of the two $|S_{11}|$. Therefore, half antennas are labeled valid in Fig.~\ref{acc0}. The boundary of valid and invalid classes in the performance space is the line with slope of $-1$ and intercept of $-5.5114$ dB because the two weights are equally set to 1. With the initial simulated dataset and the criterion of $|S_{11}|_{2.4 \  GHz}+|S_{11}|_{5.9 \  GHz} \leq -5.5114$ dB, Algorithm 1 can be applied to create a generator which is then used to obtain 100 checked potential valid designs. After simulating 100 new candidates, the results are plotted in Fig.~\ref{curr0}. 85 of them satisfy the current criterion. Also, one design satisfies $|S_{11}|<-10$ dB at both 2.4 GHz and 5.9 GHz. Its geometric model and $|S_{11}|$ curves are shown in Fig.~\ref{best1}.

Although 85\% of designs from the generator are accepted by the current criterion, majority of them do not achieve the goal of \eqref{example1} yet. Therefore, there is huge benefit to introduce the evolutionary approach here. We combine the initial dataset shown in Fig.~\ref{acc0} and the newly simulated dataset in Fig.~\ref{curr0}. Fig.~\ref{acc1} shows the performance space visualization of the combination. Then the median value of $|S_{11}|_{2.4 \  GHz}+|S_{11}|_{5.9 \  GHz}$ is selected to be the new criterion. The generative algorithm is able to train a new generator and output 100 new candidates. The simulation results are shown in Fig.~\ref{curr1}, and the acceptance rate drops to 50\% sounds causal. Then we continue the third evolution and take similar stpdf: 1) combining the dataset of Fig.~\ref{acc1} and Fig.~\ref{curr1}; 2) updating the criterion to $|S_{11}|_{2.4 \  GHz}+|S_{11}|_{5.9 \  GHz} \leq -10.5983$ dB which is the 30th percentile of the combined dataset of Fig.~\ref{acc2}; 3) training a neural network discriminator and a support vector classifier; 4) using the neural network discriminator to train a generator; 5) checking the candidates by the support vector classifier; 6) simulating new candidate designs. Last, the simulation results of the 100 newly generated antennas are shown in Fig.~\ref{curr2}. The acceptance rate for the recent criterion becomes 66\%. The updated criterion of the fourth evolution becomes $-12.6256$ dB and the acceptance rate becomes 58\%. The combined dataset at the beginning of the fifth evolution is shown in Fig.~\ref{acc3}. The criterion is updated to be the 20th percentile, i.e. $-15.8813$ dB. Till now the 100 antennas are labeled as valid in this dataset. The simulation results of the new candidates are shown in Fig.~\ref{curr3}. The updated criterion of the sixth evolution becomes $-16.9619$ dB which is the 20th percentile and the acceptance rate is 65\%. Finally among the 700 simulated designs shown in Fig.~\ref{curr4}, there are 116 different designs meeting the design target in \eqref{example1}. 

\begin{figure}[!t]
\centering \includegraphics[width=1 \columnwidth]{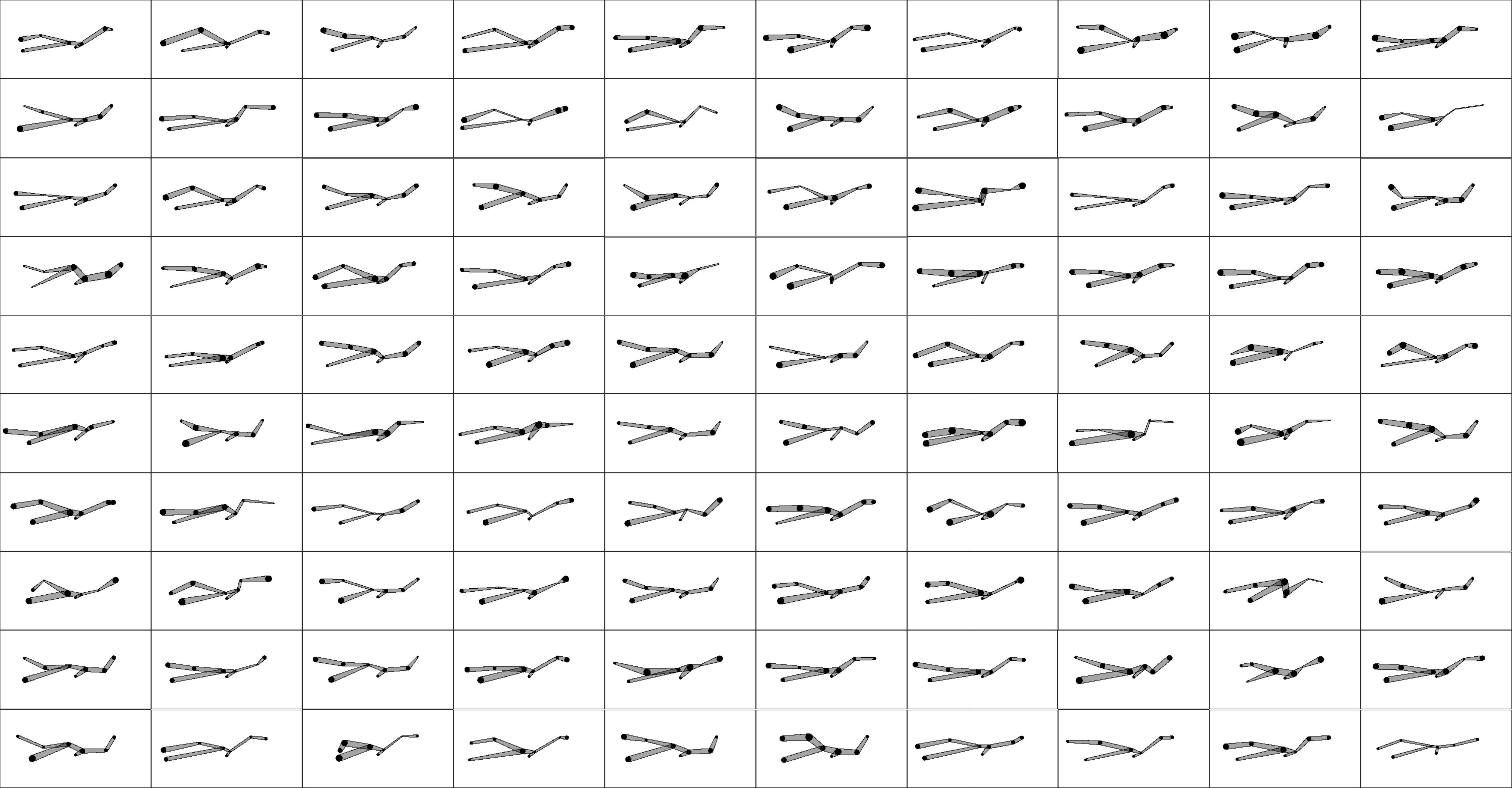}\\
  \caption{Overview of the 100 generated antennas from the most updated generator that passed the SVC in the first example.}\label{antennas_generate100}
\end{figure}

\begin{figure}[!t]
\centering \includegraphics[width=0.9 \columnwidth]{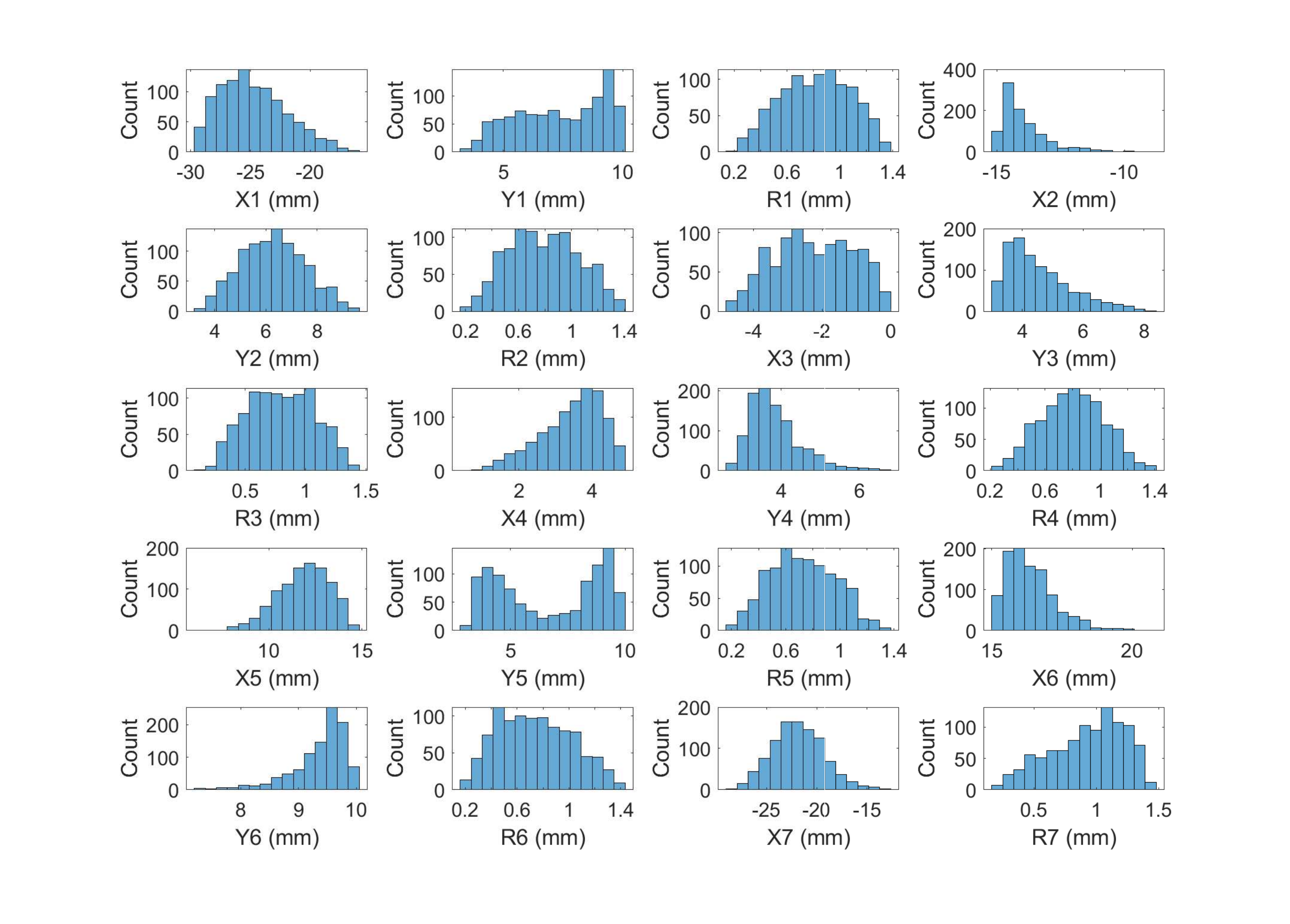}\\
  \caption{The histograms of geometric parameters based on the statistics of the 1000 generated antennas from the most updated generator in the first example. The distributions are no longer uniform.}\label{hist_final1000}
\end{figure}

Now it is the time to train a desired generator. Fig.~\ref{antennas_generate100} shows the 100 generated antennas which have already passed the SVC. The variety of the geometrical patterns is decreased from the ones in Fig.~\ref{antennas_initial100} because those newly generated antennas are supposed to be potentially valid for the design target \eqref{example1}. It is noted that the generated designs still have obvious differences, which means the most updated generator has discovered a valid space instead of a valid point with its small neighbourhood. To illustrate it, the histograms for the 20 geometric parameters are plotted in Fig.~\ref{hist_final1000}. None of them follows the uniform distribution as the initial random dataset does. For example, the parameter Y5 which represents the Y coordinate of the node 5 has two distribution peaks, one around 3 mm and the other around 9 mm. The Y coordinate of its connected node 6, Y6, is concentrated around 9.5 mm. Fig.~\ref{antennas_generate100} shows nearly half trapezoids connecting node 5 and node 6 are tilted upward with large angles, and nearly another half of them just slightly tilted.

\begin{figure}[!t]
\centering
\subfloat[]{\includegraphics[width=0.4\columnwidth]{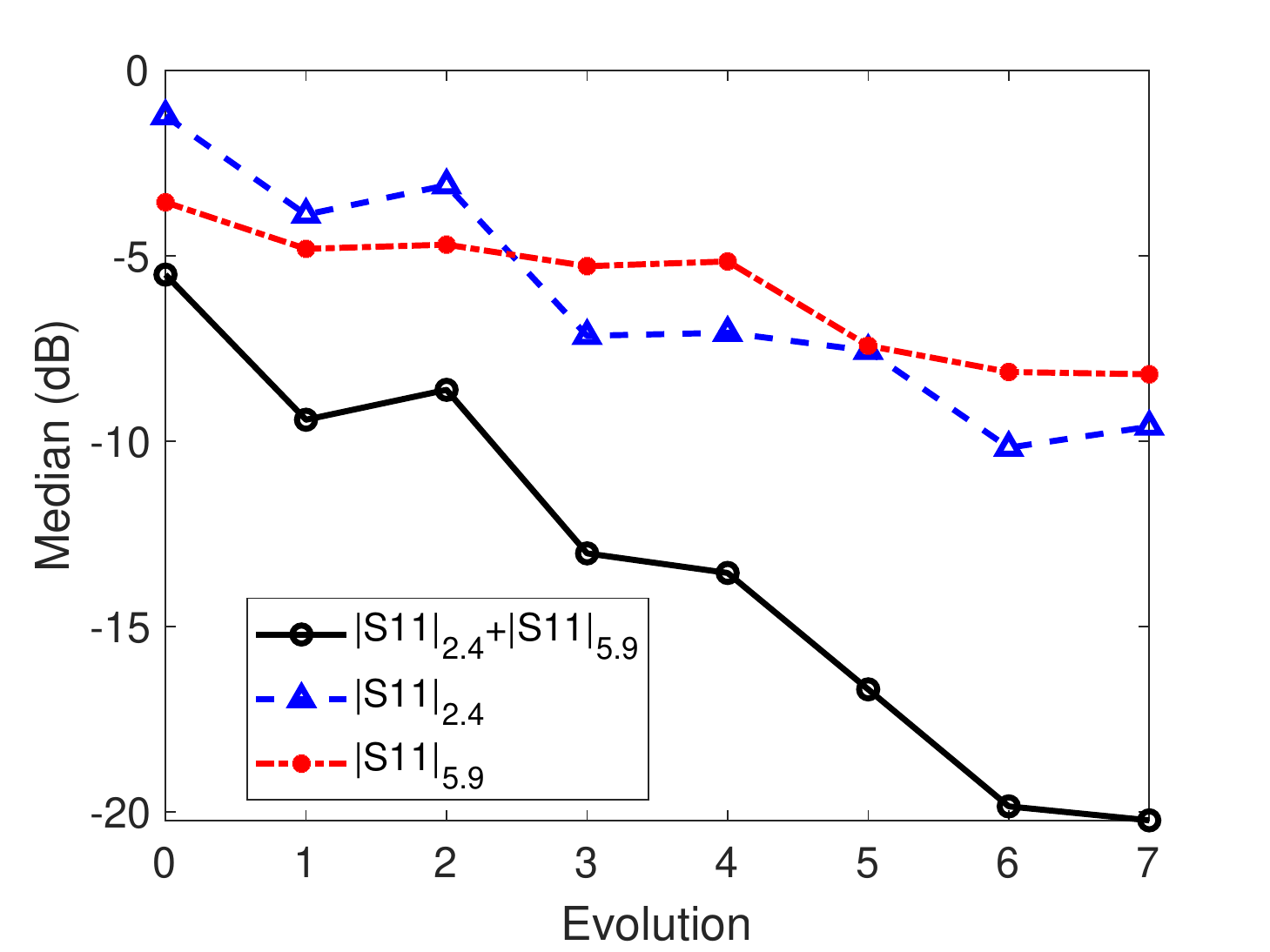}
\label{median}}
\subfloat[]{\includegraphics[width=0.4\columnwidth]{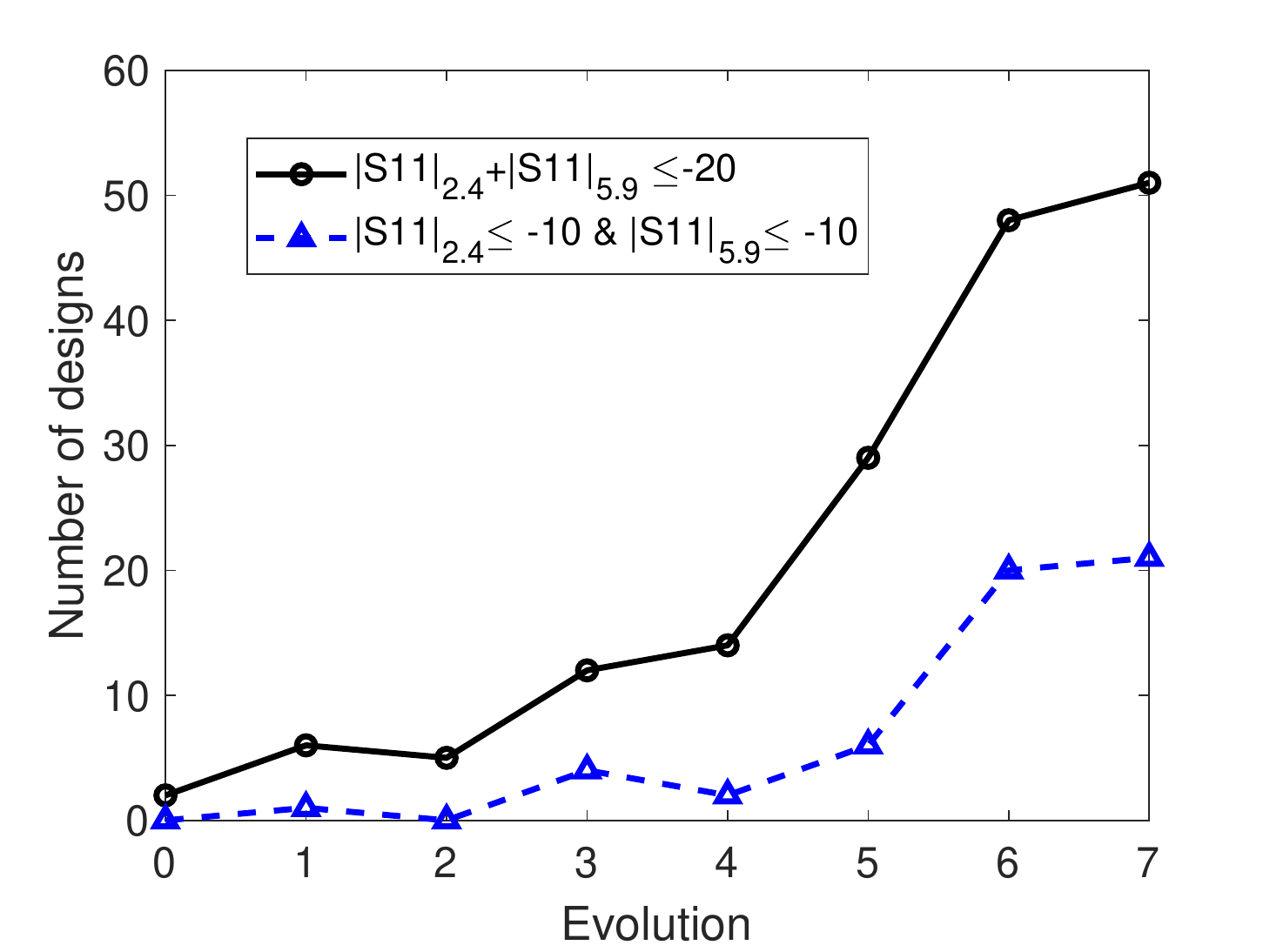}
\label{validNum}}
\caption{The statistics in each evolution of the first example. (a) The median values of $|S_{11}|$ in the generated dataset (size 100). (b) The number of antennas that achieved the goals in the generated dataset (size 100).}
\label{curves}
\end{figure}

Fig.~\ref{best} exhibits 4 valid designs obtained by using our proposed method. All of them not only meet the target in \eqref{example1}, but also achieve a stricter dual resonance requirement of $|S_{11}|<-10$ dB at both 2.4 GHz and 5.9 GHz. During the evolution of valid criterion, a simple random generator and 7 potential valid generators have been used. Fig.~\ref{curves} depicts their performance improvement. Fig.~\ref{median} illustrates the decrease of median in the 100 generated designs. At the beginning of evolution when the simple random generator starts, the median of $|S_{11}|_{2.4 \  GHz}+|S_{11}|_{5.9 \  GHz}$ is $-5.5114$ dB. Then it drops to $-20.2286$ dB when the seventh updated generator becomes available. Fig.~\ref{validNum} exhibits the number of designs that have met the design targets. Note that these numbers are also counted in the new generated 100 antennas. For the dual resonance target, it is shown that the number of valid design has been increased from 0 to 21. In other words, the most updated generator is ready to design antenna which has 20\% probability to achieve the specification of dual resonance. For the target defined in \eqref{example1}, the probability of a valid design is about 50\%. Therefore, an effective generator which supports antenna design automation has been obtained. Now once we inject a few random noise vectors into the generator, the generator will output to potential valid designs. The true valid design could be found in a few simulations. 


\subsection{Broadband Antenna Optimization}
\label{Example2}
\begin{figure}[!t]
\centering
\subfloat[]{\includegraphics[width=0.3\columnwidth]{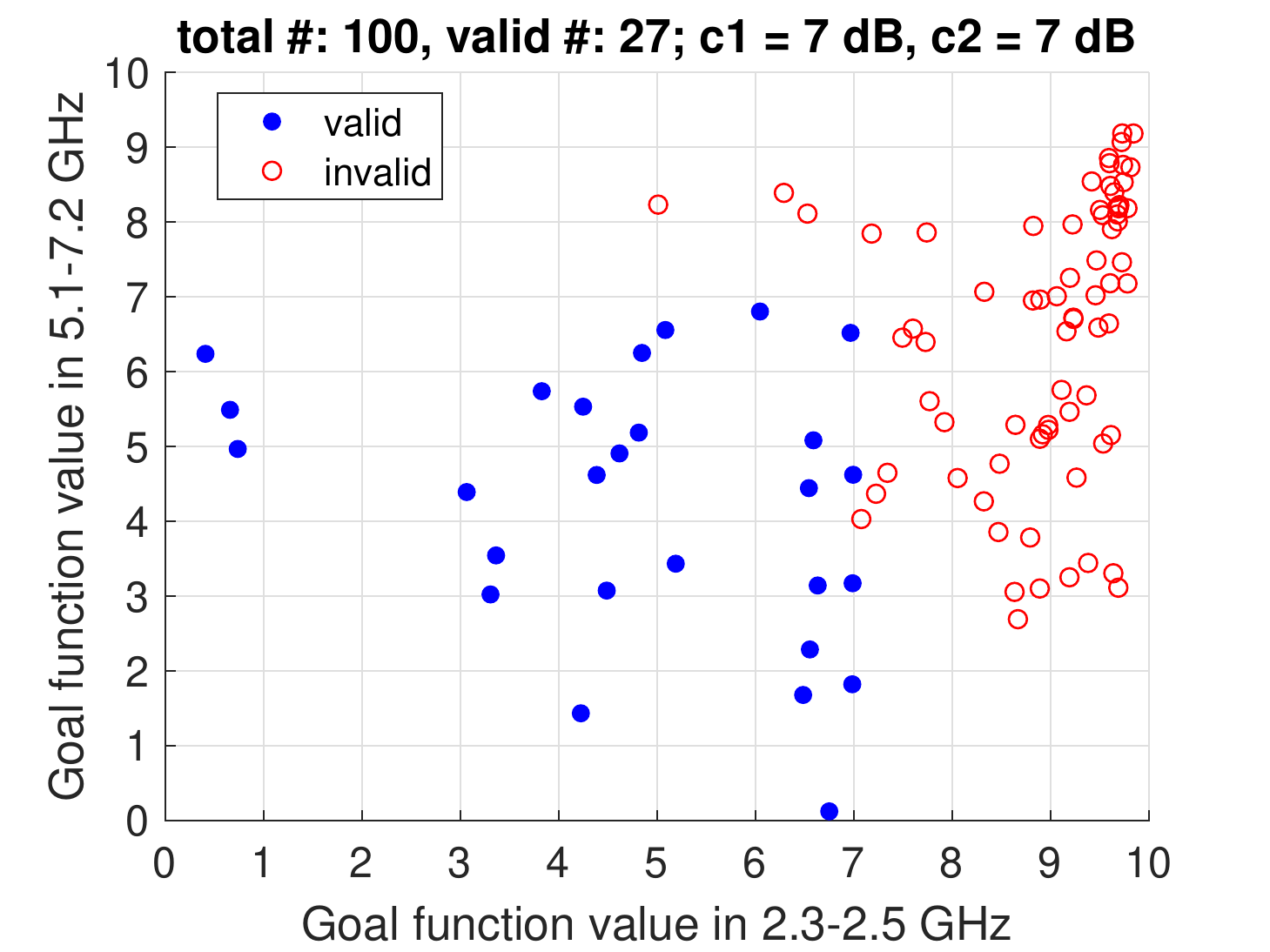}
\label{example2_acc0}}
\subfloat[]{\includegraphics[width=0.3\columnwidth]{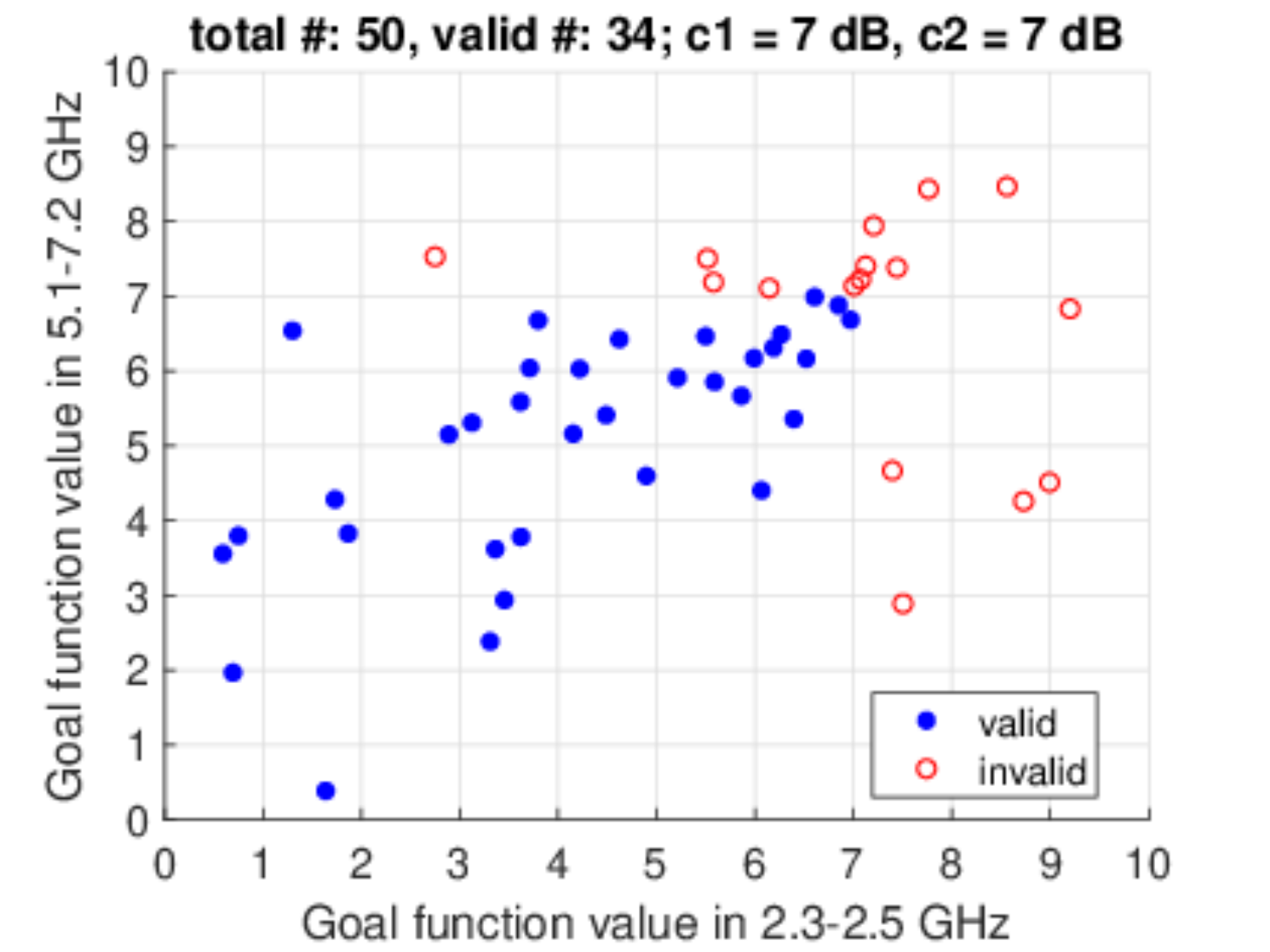}
\label{example2_curr0}}
\subfloat[]{\includegraphics[width=0.3\columnwidth]{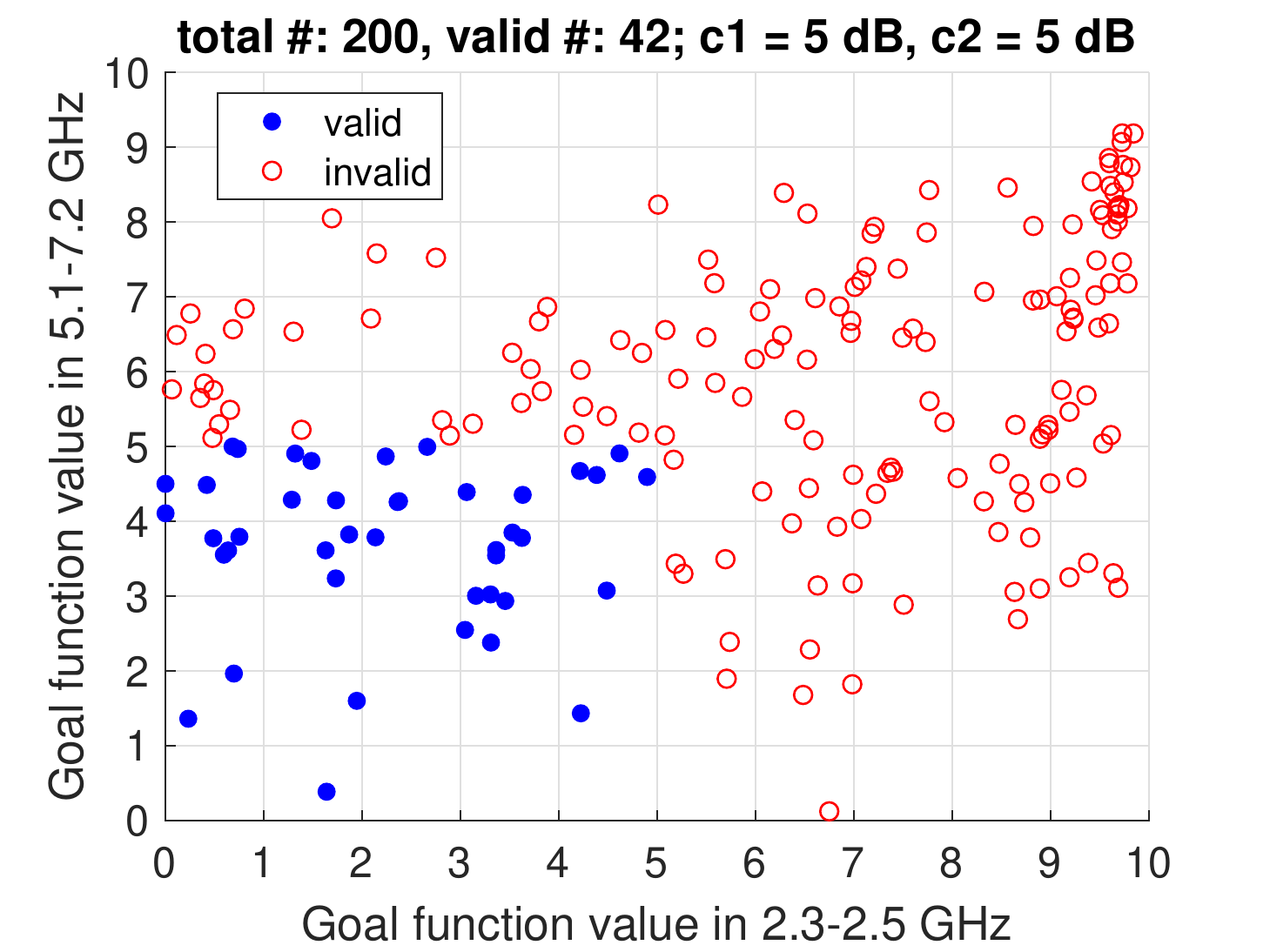}
\label{example2_acc1}}

\subfloat[]{\includegraphics[width=0.3\columnwidth]{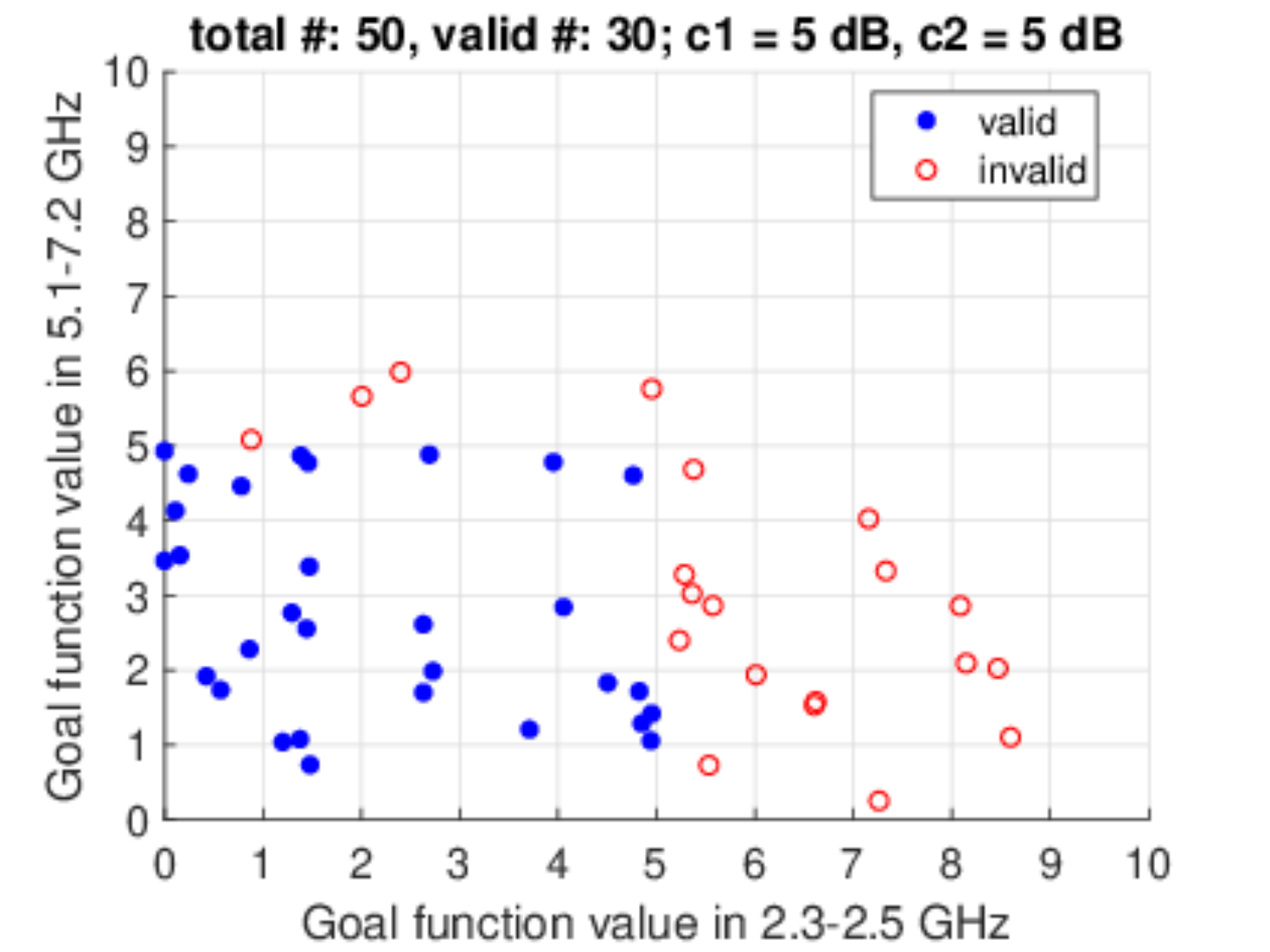}
\label{example2_curr1}}
\subfloat[]{\includegraphics[width=0.3\columnwidth]{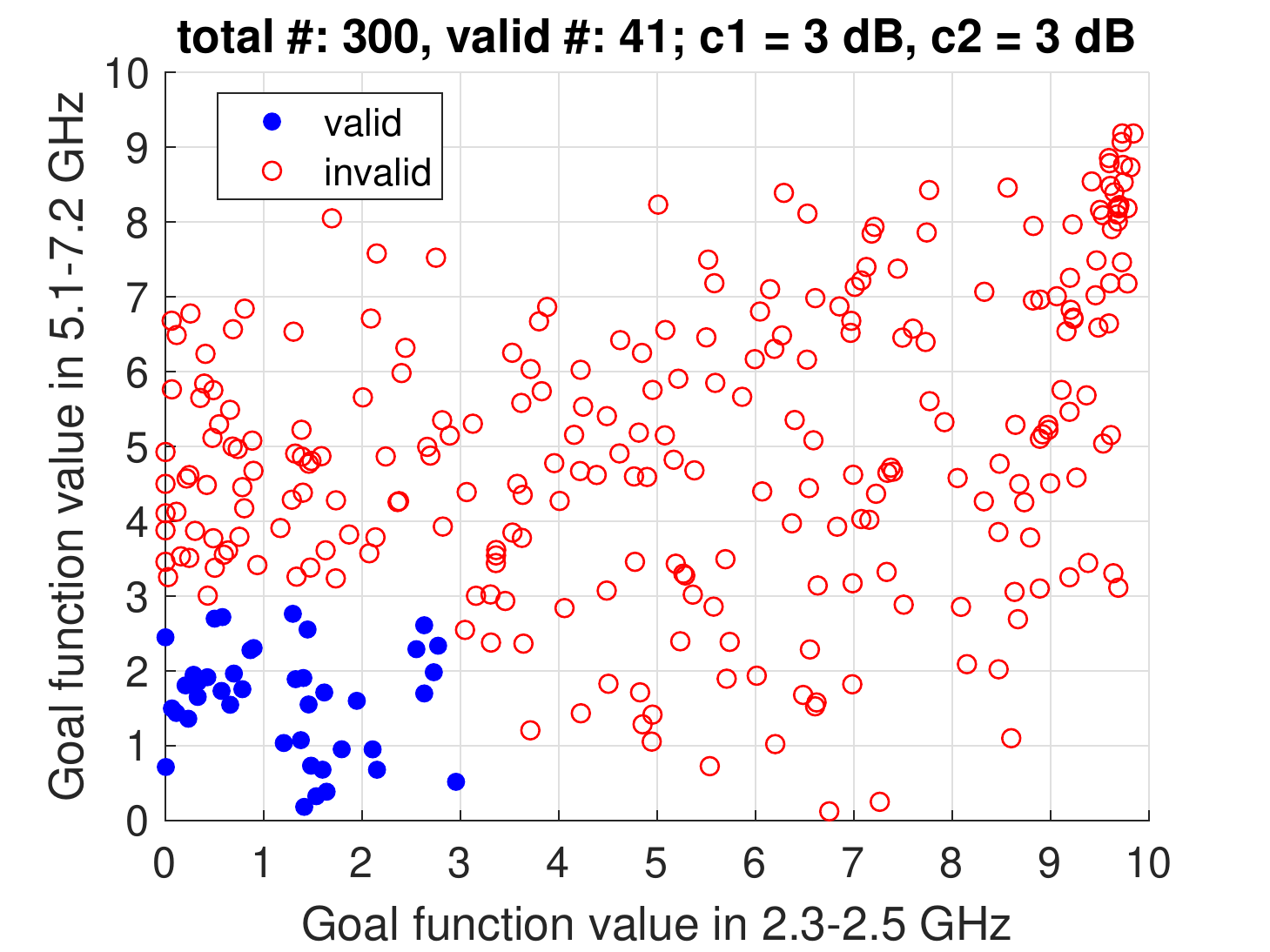}
\label{example2_acc2}}
\subfloat[]{\includegraphics[width=0.3\columnwidth]{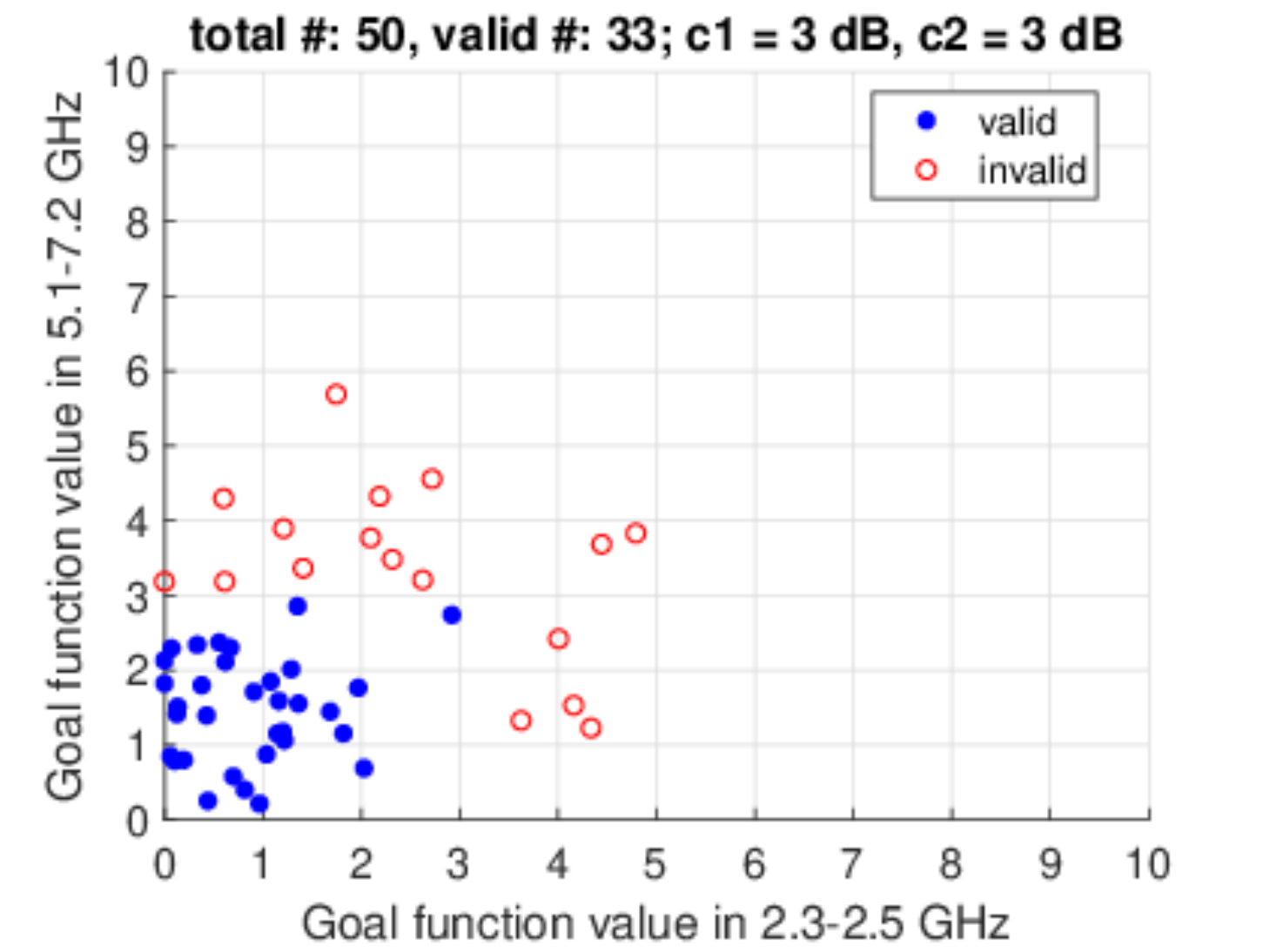}
\label{example2_curr2}}

\subfloat[]{\includegraphics[width=0.3\columnwidth]{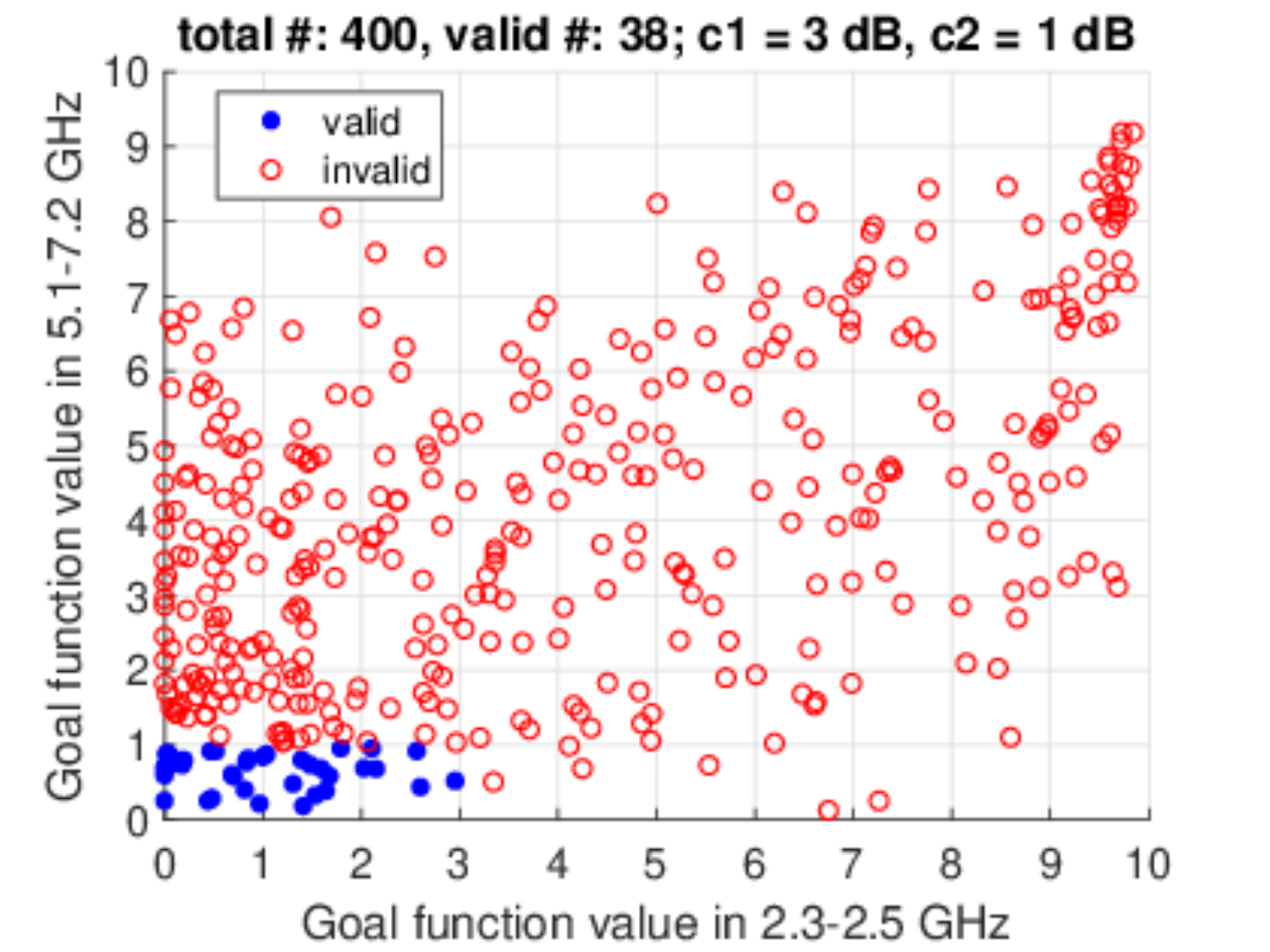}
\label{example2_acc3}}
\subfloat[]{\includegraphics[width=0.3\columnwidth]{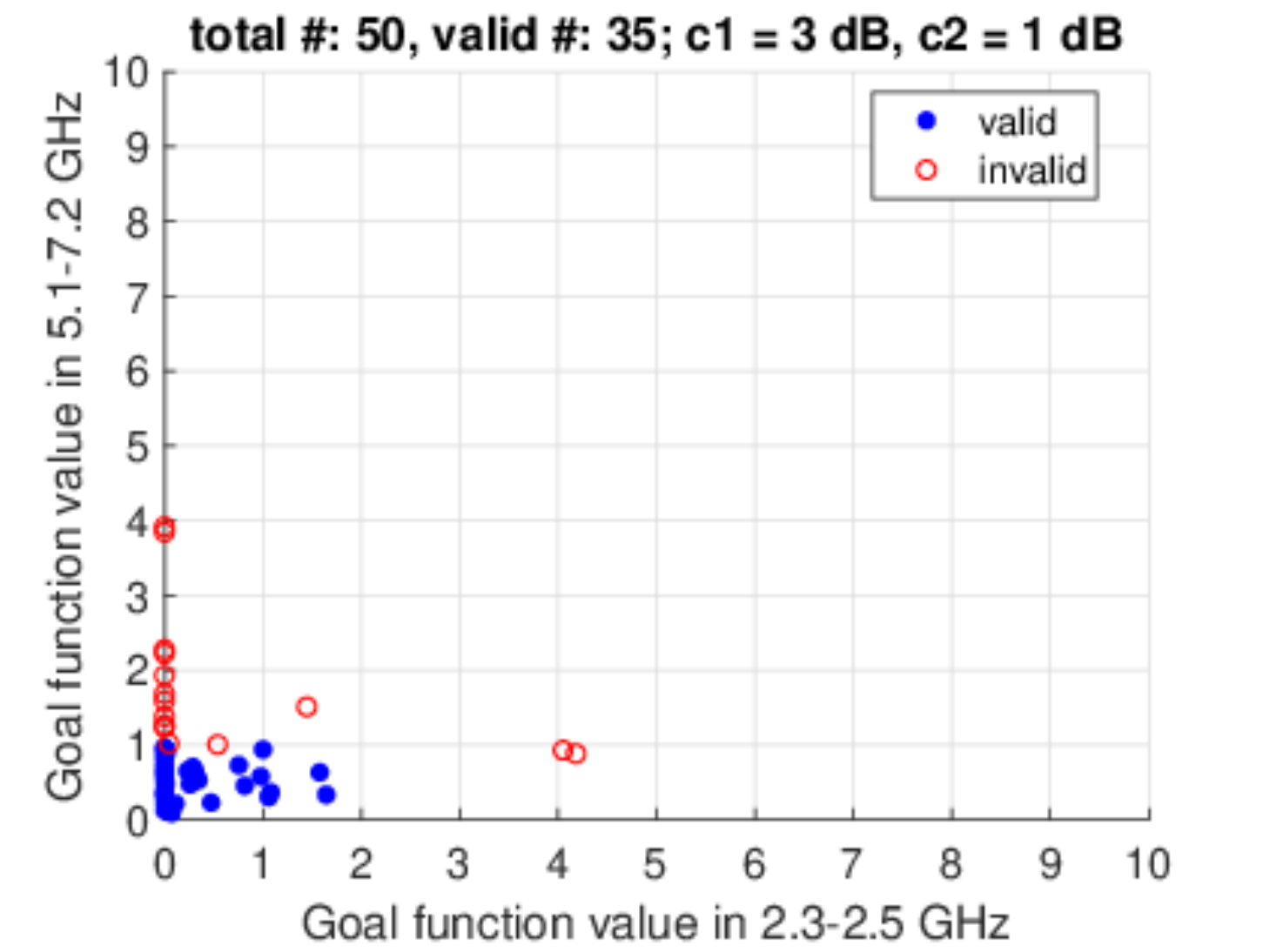}
\label{example2_curr3}}

\caption{Visualization of the dataset in the 2D performance space of the second example. Each dot represents a simulated antenna. (a) The initial dataset in the first evolution. (b) The generated dataset in the first evolution. (c) Combined dataset in the third evolution. (d) The generated dataset in the third evolution. (e) Combined dataset in the fifth evolution. (f) The generated dataset in the fifth evolution. (g) Combined dataset in the seventh evolution. (h) The generated dataset in the seventh evolution.}
\label{example2_intermediate}
\end{figure}

\begin{figure}[!t]
\centering
\subfloat[]{\includegraphics[width=0.4\columnwidth]{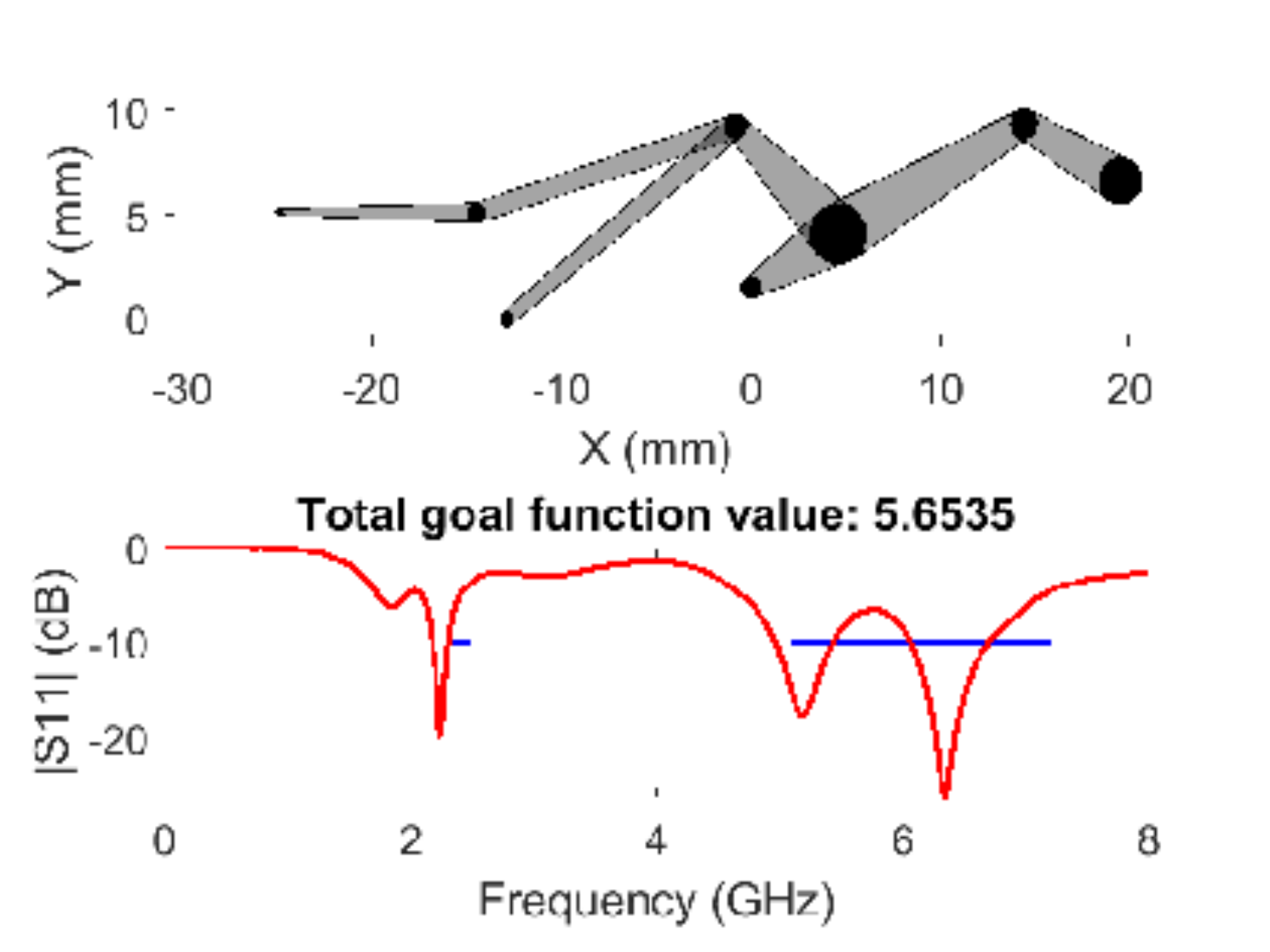}
\label{example2_best1}}
\subfloat[]{\includegraphics[width=0.4\columnwidth]{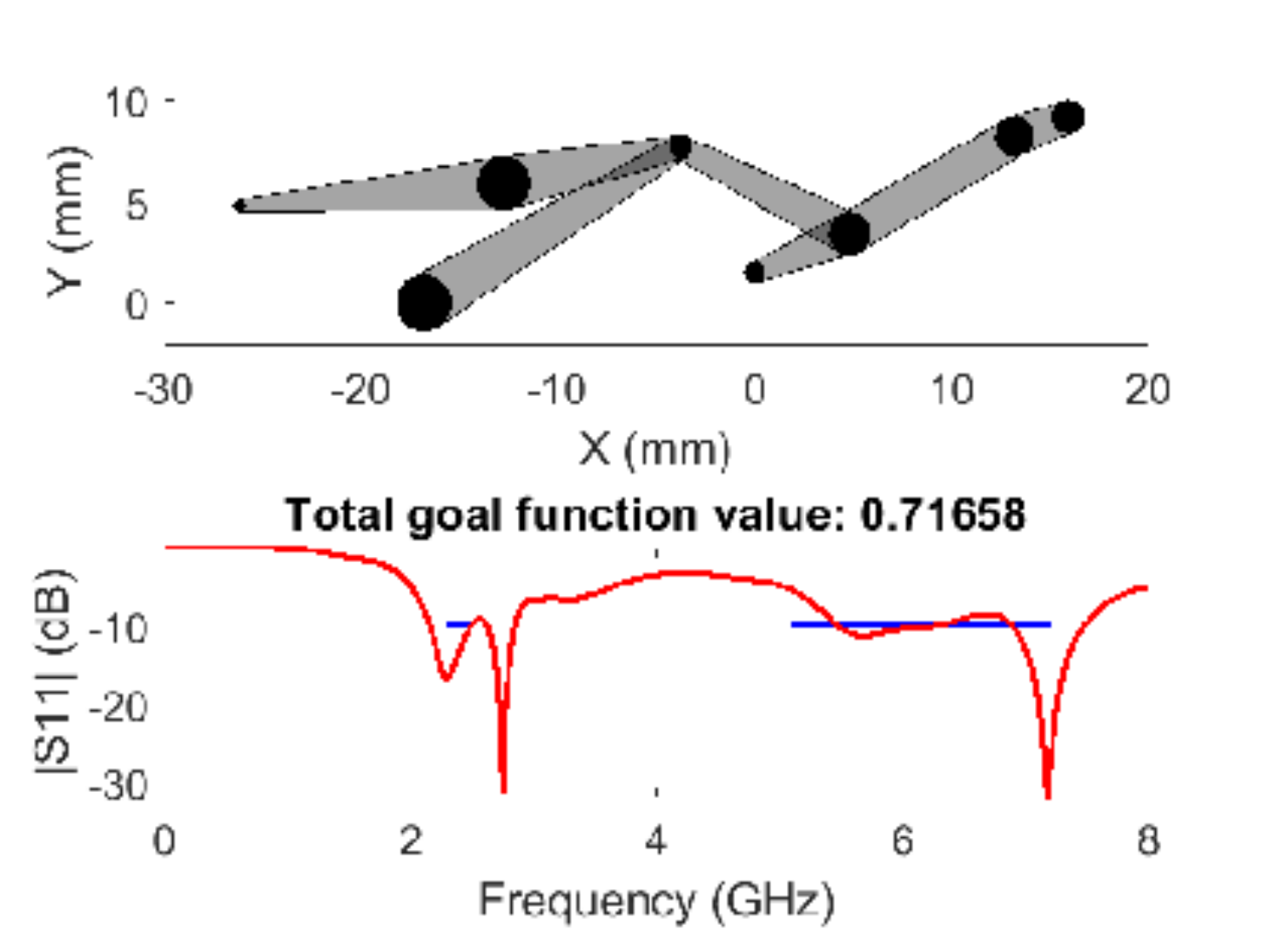}
\label{example2_best2}}

\subfloat[]{\includegraphics[width=0.4\columnwidth]{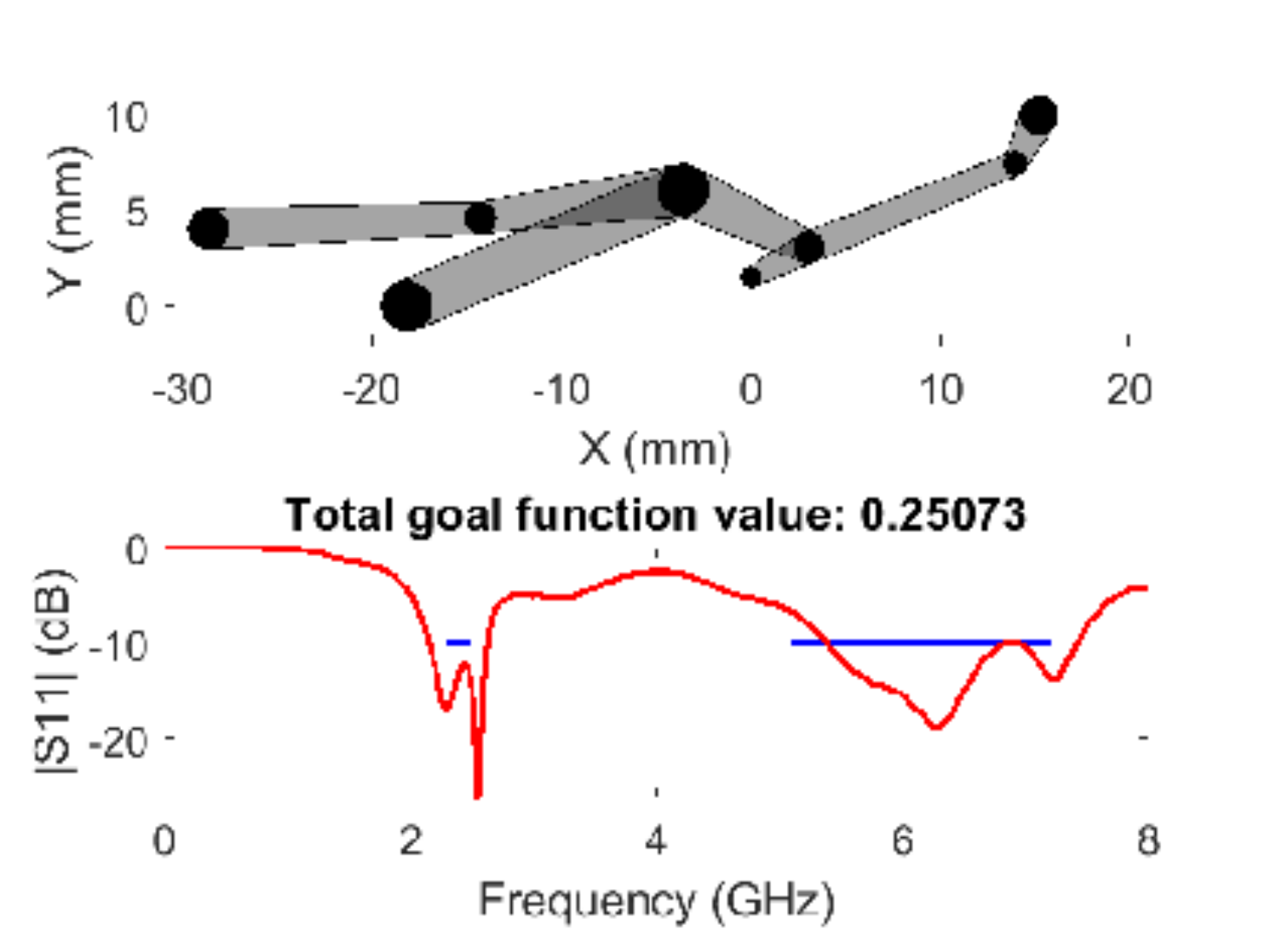}
\label{example2_best3}}
\subfloat[]{\includegraphics[width=0.4\columnwidth]{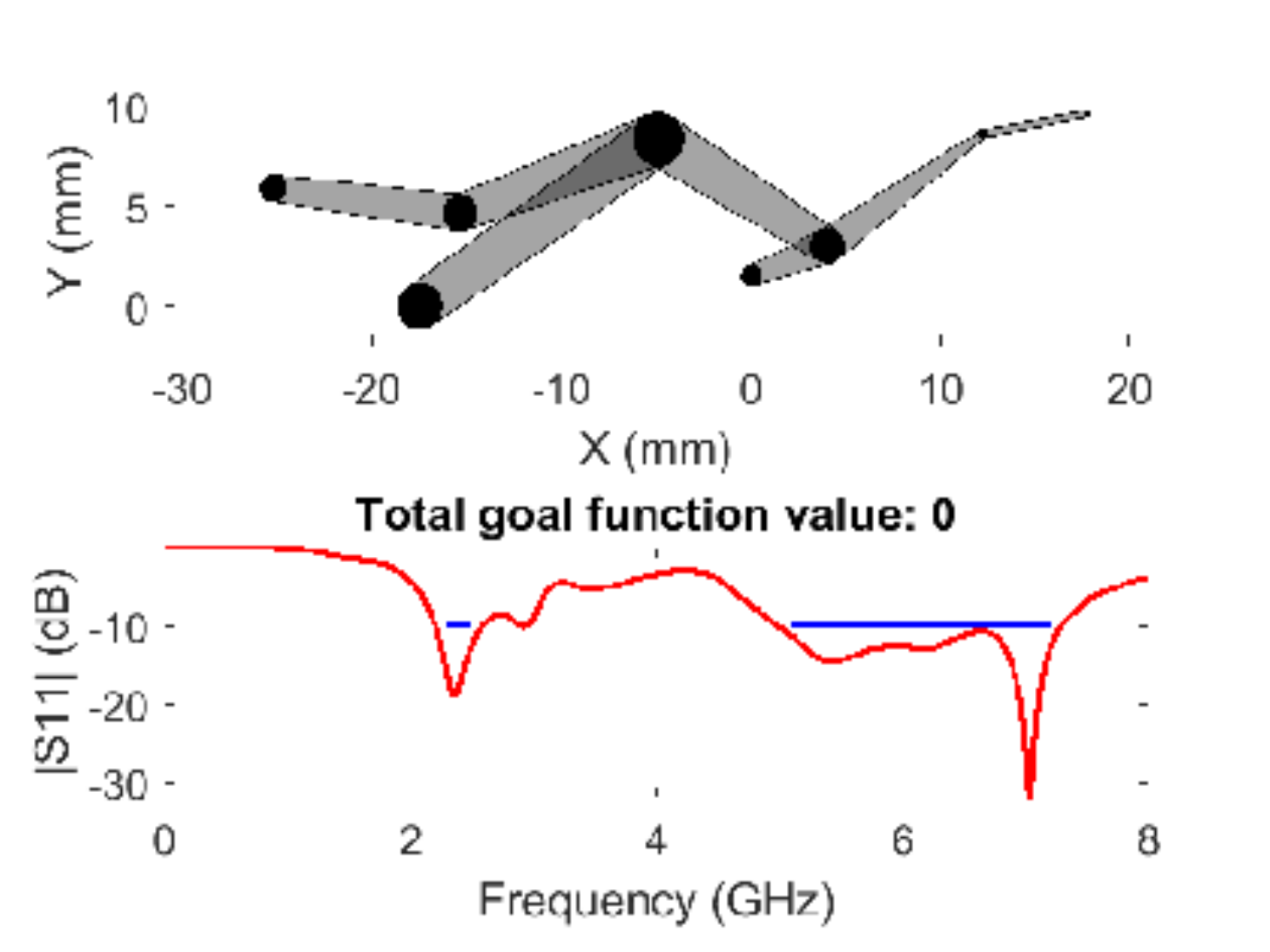}
\label{example2_best4}}

\caption{The antenna geometric model and the corresponding $S_{11}$ curve in the second example. (a) The best antenna in the initial random dataset. (b) The best generated antenna in the fourth evolution. (c) The best generated antenna in the sixth evolution. (d) The best generated antenna in the eighth evolution.}
\label{example2_best}
\end{figure}

Now let's investigate the second example on how to broaden the antenna bandwidth. Using the same antenna geometric modeling scheme as illustrated in Fig.~\ref{connectingNodes}, we specify more realistic design targets for wider bandwidth, i.e. $|S_{11}| \leq -10$ dB in the frequency ranges of 2.3-2.5 GHz and 5.1-7.2 GHz. The desired bandwidth covers the 2.4/5 GHz WiFi as well as Sub-6 Cellular bands. For this example, we define a total target function:
\begin{eqnarray} \label{example2}
F=f(|S_{11}|_{[2.3,2.5]\  GHz}+10)) + f(|S_{11}|_{[5.1,7.2]\  GHz}+10)),
\end{eqnarray}
which is the sum of two target functions. The target function $f$ is defined by
\begin{eqnarray} \label{example2_}
f(X) = \frac{1}{N}\sum_{i=1}^{N}max(X_i,0).
\end{eqnarray}
The target function value in a frequency range represents the average difference between $|S_{11}|$ and the target value of $-10$ dB. If the $|S_{11}| \leq -10$ dB, the difference will be 0. With such definition, we can visualize the simulation results of the initial 100 random dataset in Fig.~\ref{example2_acc0}. The criterion for splitting the dataset follows \eqref{criterion1}, where $p_1$ and $p_2$ are the target function values in 2.3-2.5 GHz and 5.1-7.2 GHz respectively. Both thresholds are 7 dB. Only 27 antennas are marked valid out of the 100 random antennas. The best one with the smallest total target function value is shown in Fig.~\ref{example2_best1}, where the sum of average differences of $|S_{11}|$ to $-10$ dB in the interested frequency range is 5.6535 dB. 

The evolutionary approach with the generative algorithm is proposed to find a design whose total target function value equals zero. The simulation results of the designs from the first potential valid generator are shown in Fig.~\ref{example2_curr0}. Of the 50 candidates, 34 are valid for the initial criterion. When the first evolution is done, we combine the initial 100 dataset and the new 50 dataset to get a dataset with 150 simulated antennas. In the second evolution, we update thresholds to 6 dB. There are 32 valid antennas in the 150 dataset. In the 50 simulated candidates, the valid number is 37. In the third evolution, Fig.~\ref{example2_acc1} shows the combined dataset where the criterion is updated to 5 dB. 42 out of 200 antennas are labeled valid. We update the discriminators and the generator, and there are 30 out of 50 valid as shown in Fig.~\ref{example2_curr1}. In the fourth evolution, the criterion becomes 4 dB, 36 out of 250 antennas are valid. The updated generator outputs to 39 valid designs. Fig.~\ref{example2_best2} shows the best design among the 50 generated candidates. The total target function value has been decreased to 0.71658 dB. In the fifth evolution, the criterion is 3 dB. 41 out of 300 are valid as shown in Fig.~\ref{example2_acc2}. Among the generated 50 candidates, there are 33 valid designs as shown in Fig.~\ref{example2_curr2}. In the sixth evolution, the criterion becomes 3 dB for 2.3-2.5 GHz, 2 dB for 5.1-7.2 GHz. 55 out of 350 are valid. The reason why we choose the criterion in this way is because from our observation it is harder to reduce the target function value in 5.1-7.2 GHz than that in 2.3-2.5 GHz. Out of the generated 50 candidates, there are 34 valid designs. Fig.~\ref{example2_best3} shows the best design in the 50 generated candidates. The total target function value has decreased to 0.25073 dB. In the seventh evolution, the criterion becomes 3 dB for 2.3-2.5 GHz, 1 dB for 5.1-7.2 GHz. 38 out of 400 are valid as shown in Fig.~\ref{example2_acc3}. Among the generated 50 candidates, there are 35 valid designs as shown in  Fig.~\ref{example2_curr3}. The generated smallest total goal function value is 0.1203 dB. In the eighth evolution, the criterion is 1 dB for both frequency ranges. 51 out of 450 are valid for training the discriminators. Out of the generated 50 candidates, there are 38 valid designs. One of them achieves the design target, i.e. its total goal function value is 0. Fig.~\ref{example2_best4} shows the best design of the generated candidates. Till now, we terminate the evolution process because a solution has been identified.

For this example with 20 degrees of freedom, the design target is finally achieved with only 100 initial random antennas. It is possible that the iteration cannot terminate because the target cannot be achieved. For such a problem, we may need more initial random antennas to let us interpolate the antenna geometric models in a wider space.

\subsection{Comparison of Optimizers}
For comparison with previous optimization methods, we select a few existing optimizers embedded in the CST Studio \cite{CST}. We use the same design targets and the exactly same model scheme in Fig.~\ref{connectingNodes} for each optimizer. The minimum numbers of simulations used to find the solution are listed in Table II. It shows the efficiency of each optimizer. For the dual band antenna with broad bandwidth at both resonances, our proposed method is in par with Trust Region Framework and much better than the others especially the widely used Genetic Algorithm and Particle Swarm Optimization. When there is no strict bandwidth requirement, our proposed method is better than Trust Region Framework and in par with Genetic Algorithm and CMA Evolution Strategy. The model architecture and training parameters are recorded in the Appendix for reference. 

Since in this paper, the only metrics used to compare different optimizers is the minimum number of simulations, we also want to point out that our proposed method is indeed friendly for parallel simulation. In each evolution, the generated antennas awaiting for simulation are independent from each other. Any number of simulations can be run simultaneously. It is known that CST can support parallel computation with limitations. For example, the Trust Region Framework embedded in CST only can run maximum 20 simulations at a time because it needs to decide next candidates based on the partial derivatives of each design parameters. Therefore, the proposed method shines a light on massive parallel computation for future black-box antenna design and optimization. 

\begin{table}[h!]
\caption{Comparison of Optimization Algorithms}
\label{table:2}
\centering
\begin{tabular}{c c c} 
 \hline
 
 \hline
{Number of Simulations} & {$|S_{11}|<-10$ dB @2.4, 5.9 GHz} & {$|S_{11}|<-10$ dB in 2.3-2.5, 5.1-7.2 GHz}  \\  
 \hline
 {Proposed Method} & 139 & 472  \\ 
 \hline
 {Trust Region Framework} & 232 & 430  \\ 
 \hline
 {Nelder Simplex Algorithm} & 88 & 617  \\ 
 \hline
 {CMA Evolution Strategy} & 144 & 749  \\ 
 \hline
 {Genetic Algorithm} & 108 & >1000  \\ 
 \hline 
 {Particle Swarm Optimization} & 43 & >1000 \\ 
 \hline 
 
 \hline
 \end{tabular}
\end{table}

\section{Conclusions}
\label{secConc}
Apparently, in the above two examples of dual resonance antenna design and broadband optimization, we did not mention much antenna domain terminologies such as surface current distribution, Smith chart and etc. The reason is that we are aiming for using artificial intelligent and machine learning algorithms to design black-box antennas in order to free engineering efforts, not only limited to optimizing a few geometric parameters. In this paper, we introduced the connecting node modeling scheme which has flexibility of being configured and reconfigured to antenna geometric shapes, and proposed a machine learning method which is able to assist us to achieve the goal. 

\appendix
\section{Appendix}
The neural networks in the two examples are similar. Table II shows the model architectures of the discriminator and generator. The feature map numbers marked by star means they are adjustable. Table III gives the exact feature maps in each evolution for the first example. Table IV records the training parameters used in PyTorch. 
\label{FirstAppendix}
\begin{table}[h!]
\caption{Model architectures of the discriminator and generator}
\label{table:a1}
\centering
\begin{tabular}{c c c c c} 
 \hline
Operation & Feature maps & Batch norm? & Bias? & Nonlinearity\\ 
 \hline
 $D(x)$ -- $20$ &  &  &  & \\
 \hline
 Linear &  64* & No & Yes & ReLU \\
 \hline
 Linear &  128* & No & Yes & ReLU \\
 \hline
 Linear &  256* & No & Yes & ReLU \\
 \hline
 Linear &  1 & No & Yes & Sigmoid \\
 \hline
  $G_x(z)$ -- $20 $ &  &  &  & \\
 \hline
 Linear &  128 & No & No & Leaky ReLU \\
 \hline
 Linear &  256 & Yes & No & Leaky ReLU \\
 \hline
 Linear &  512 & Yes & No & Leaky ReLU \\
 \hline
 Linear & 20 & No & No & Sigmoid \\ 
 \hline
 \end{tabular}
 \end{table}

\begin{table}[h!]
\caption{Discriminator Information At Each Evolution}
\label{table:a2}
\centering
\begin{tabular}{c c c c c} 
 \hline
Evolution & 1st layer & 2nd layer & 3rd layer & Test accuracy\\ 
 \hline
 1 &  64 & 128 & 256 & 85\%  \\
 \hline
 2 &  128 & 128 & 256 & 76\%  \\
 \hline
 3 &  256 & 512 & 512 & 82\%  \\
 \hline
 4 &  256 & 512 & 256 & 79\%  \\
 \hline
 5 &  256 & 512 & 512 & 82\%  \\
 \hline
 6 &  256 & 256 & 1024 & 84\%  \\
 \hline
 7 &  256 & 256 & 1024 & 88\%  \\
 \hline
 \end{tabular}
 \end{table}

 \begin{table}[h!]
 \caption{Hyperparameters for training the discriminator and generator}
\label{table:a3}
\centering
 \begin{tabular}{c c} 
 \hline
 Discriminator Optimizer & Adam\ ($lr = [10^{-6},10^{-5}], decay=0.05$) \\
 Generator Optimizer & SGD\ ($lr = [10^{-5},10^{-4}]$) \\
 Discriminator batch size & 1 \\
 Generator batch size & 500 \\
 Discriminator epoch numbers & $\leq 1000$ \\
 Generator epoch numbers &  $\leq 30000$ \\
 Leaky ReLU slope & 0.2 \\
 Noise distribution & Uniform\ (-1,1) \\
 Initialization of weight & Kaiming uniform\\
 Initialization of bias & Constant 0\\
 \hline
\end{tabular}
\end{table}

\bibliographystyle{unsrtnat}
\bibliography{references}  

\begin{thebibliography}{28}
\providecommand{\natexlab}[1]{#1}
\providecommand{\url}[1]{\texttt{#1}}
\expandafter\ifx\csname urlstyle\endcsname\relax
  \providecommand{\doi}[1]{doi: #1}\else
  \providecommand{\doi}{doi: \begingroup \urlstyle{rm}\Url}\fi

\bibitem[Behdad and Sarabandi(2004)]{Behdad2004BandwidthEA}
Nader Behdad and Kamal Sarabandi.
\newblock Bandwidth enhancement and further size reduction of a class of
  miniaturized slot antennas.
\newblock \emph{IEEE Transactions on Antennas and Propagation}, 52:\penalty0
  1928--1935, 2004.

\bibitem[Wong et~al.(2008)Wong, Mak, and Luk]{Wong2008WidebandSB}
Hang Wong, Ka~Ming Mak, and Kwai~Man Luk.
\newblock Wideband shorted bowtie patch antenna with electric dipole.
\newblock \emph{IEEE Transactions on Antennas and Propagation}, 56:\penalty0
  2098--2101, 2008.

\bibitem[Qing et~al.(2010)Qing, Goh, and Chen]{Qing2010ABU}
Xianming Qing, Chean~Khan Goh, and Zhi~Ning Chen.
\newblock A broadband {UHF} near-field {RFID} antenna.
\newblock \emph{IEEE Transactions on Antennas and Propagation}, 58:\penalty0
  3829--3838, 2010.

\bibitem[Pazin and Leviatan(2011)]{Pazin2011InvertedFLA}
L.~Z. Pazin and Yehuda Leviatan.
\newblock {Inverted-F} laptop antenna with enhanced bandwidth for {Wi-Fi/WiMAX}
  applications.
\newblock \emph{IEEE Transactions on Antennas and Propagation}, 59:\penalty0
  1065--1068, 2011.

\bibitem[chun Tang et~al.(2016)chun Tang, Shi, and
  Ziolkowski]{Tang2016PlanarUA}
Ming chun Tang, Ting Shi, and Richard~W. Ziolkowski.
\newblock Planar ultrawideband antennas with improved realized gain
  performance.
\newblock \emph{IEEE Transactions on Antennas and Propagation}, 64:\penalty0
  61--69, 2016.

\bibitem[Su et~al.(2019)Su, Lee, and Hsiao]{Su2019CompactTS}
Saou-Wen Su, Cheng-Tse Lee, and Ya-Wen Hsiao.
\newblock Compact two-inverted-{F}-antenna system with highly integrated
  pi-shaped decoupling structure.
\newblock \emph{IEEE Transactions on Antennas and Propagation}, 67:\penalty0
  6182--6186, 2019.

\bibitem[Alexandrov et~al.(1997)Alexandrov, Dennis, Lewis, and
  Torczon]{Alexandrov1997ATF}
N.~M. Alexandrov, John~E. Dennis, R.~Michael Lewis, and Virginia Torczon.
\newblock A trust-region framework for managing the use of approximation models
  in optimization.
\newblock \emph{Structural optimization}, 15:\penalty0 16--23, 1997.

\bibitem[Dennis and Woods(1985)]{Dennis1985OptimizationOM}
John~E. Dennis and Daniel~J. Woods.
\newblock Optimization on microcomputers: The nelder-mead simplex algorithm.
\newblock 1985.

\bibitem[Hansen(2016)]{Hansen2016TheCE}
Nikolaus Hansen.
\newblock The {CMA} evolution strategy: A tutorial.
\newblock \emph{ArXiv}, abs/1604.00772, 2016.

\bibitem[Goldberg(1988)]{Goldberg1988GeneticAI}
David~E. Goldberg.
\newblock Genetic algorithms in search optimization and machine learning.
\newblock 1988.

\bibitem[Kennedy and Eberhart(1995)]{Kennedy1995ParticleSO}
James Kennedy and Russell~C. Eberhart.
\newblock Particle swarm optimization.
\newblock \emph{Proceedings of ICNN'95 - International Conference on Neural
  Networks}, 4:\penalty0 1942--1948 vol.4, 1995.

\bibitem[Burrascano et~al.(1999)Burrascano, Fiori, and
  Mongiardo]{Burrascano1999ARO}
Pietro Burrascano, Simone G.~O. Fiori, and Mauro Mongiardo.
\newblock A review of artificial neural networks applications in microwave
  computer-aided design.
\newblock 1999.

\bibitem[jun Zhang et~al.(2003)jun Zhang, Gupta, and
  Devabhaktuni]{Zhang2003ArtificialNN}
Qi~jun Zhang, K.~C. Gupta, and Vijay~Kumar Devabhaktuni.
\newblock Artificial neural networks for {RF} and microwave design - from
  theory to practice.
\newblock \emph{IEEE Transactions on Microwave Theory and Techniques},
  51:\penalty0 1339--1350, 2003.

\bibitem[Zhu et~al.(2007)Zhu, Bandler, Nikolova, and
  Koziel]{ZHU2007TAP_Special_issue}
Jiang Zhu, John~W. Bandler, Natalia~K. Nikolova, and Slawomir Koziel.
\newblock Antenna optimization through space mapping.
\newblock \emph{IEEE Transactions on Antennas and Propagation}, 55:\penalty0
  651--658, 2007.

\bibitem[Pastorino and Randazzo(2005)]{Pastorino2005ASA}
Matteo Pastorino and Andrea Randazzo.
\newblock A smart antenna system for direction of arrival estimation based on a
  support vector regression.
\newblock \emph{IEEE Transactions on Antennas and Propagation}, 53:\penalty0
  2161--2168, 2005.

\bibitem[Xiao et~al.(2018)Xiao, Shao, long Jin, and
  Wang]{Xiao2018MultiparameterMW}
Li-Ye Xiao, Wei Shao, Fu~long Jin, and Bing-Zhong Wang.
\newblock Multiparameter modeling with {ANN} for antenna design.
\newblock \emph{IEEE Transactions on Antennas and Propagation}, 66:\penalty0
  3718--3723, 2018.

\bibitem[Cui et~al.(2020)Cui, Zhang, Zhang, and Liu]{Cui2020AME}
Liangze Cui, Yao Zhang, Runren Zhang, and Qing~Huo Liu.
\newblock A modified efficient {KNN} method for antenna optimization and
  design.
\newblock \emph{IEEE Transactions on Antennas and Propagation}, 68:\penalty0
  6858--6866, 2020.

\bibitem[Xiao et~al.(2021)Xiao, Shao, Jin, Wang, and Liu]{9395365}
Li-Ye Xiao, Wei Shao, Fu-Long Jin, Bing-Zhong Wang, and Qing~Huo Liu.
\newblock Inverse artificial neural network for multiobjective antenna design.
\newblock \emph{IEEE Transactions on Antennas and Propagation}, 69\penalty0
  (10):\penalty0 6651--6659, 2021.
\newblock \doi{10.1109/TAP.2021.3069543}.

\bibitem[Sharma et~al.(2020)Sharma, Zhang, and Xin]{Sharma2020MachineLT}
Yashika Sharma, Hao~Helen Zhang, and Hao Xin.
\newblock Machine learning techniques for optimizing design of double
  {T}-shaped monopole antenna.
\newblock \emph{IEEE Transactions on Antennas and Propagation}, 68:\penalty0
  5658--5663, 2020.

\bibitem[Wu et~al.(2020)Wu, Wang, and Hong]{Wu2020MultistageCM}
Qi~Wu, Haiming Wang, and Wei Hong.
\newblock Multistage collaborative machine learning and its application to
  antenna modeling and optimization.
\newblock \emph{IEEE Transactions on Antennas and Propagation}, 68:\penalty0
  3397--3409, 2020.

\bibitem[Nan et~al.(2021)Nan, Xie, Gao, Song, and Yang]{Nan2021DesignOU}
Jingchang Nan, Huan Xie, Mingming Gao, Yang Song, and Wendong Yang.
\newblock Design of {UWB} antenna based on improved deep belief network and
  extreme learning machine surrogate models.
\newblock \emph{IEEE Access}, 9:\penalty0 126541--126549, 2021.

\bibitem[Davoli et~al.(2021)Davoli, Guerzoni, and Vitetta]{Davoli2021MachineLA}
Alessandro Davoli, Giorgio Guerzoni, and Giorgio~Matteo Vitetta.
\newblock Machine learning and deep learning techniques for colocated {MIMO}
  radars: A tutorial overview.
\newblock \emph{IEEE Access}, 9:\penalty0 33704--33755, 2021.

\bibitem[Zhou et~al.(2021)Zhou, Yang, Si, Kang, Li, Wang, and
  Zhang]{Zhou2021ATP}
Jinzhu Zhou, Zhanbiao Yang, Yu~Si, Le~Kang, Haitao Li, Mei Wang, and Zhiya
  Zhang.
\newblock A trust-region parallel bayesian optimization method for
  simulation-driven antenna design.
\newblock \emph{IEEE Transactions on Antennas and Propagation}, 69:\penalty0
  3966--3981, 2021.

\bibitem[Naseri and Hum(2021)]{Naseri2021AGM}
Parinaz Naseri and Sean~Victor Hum.
\newblock A generative machine learning-based approach for inverse design of
  multilayer metasurfaces.
\newblock \emph{IEEE Transactions on Antennas and Propagation}, 69:\penalty0
  5725--5739, 2021.

\bibitem[Koziel et~al.(2021)Koziel, Çalik, Mahouti, and
  Belen]{Koziel2021AccurateMO}
Slawomir Koziel, Nurullah Çalik, Peyman Mahouti, and Mehmet~Ali Belen.
\newblock Accurate modeling of antenna structures by means of domain
  confinement and pyramidal deep neural networks.
\newblock \emph{IEEE Transactions on Antennas and Propagation}, 2021.

\bibitem[Abdullah and Koziel(2022)]{Abdullah2022SupervisedLearningBasedDO}
Muhammad Abdullah and Slawomir Koziel.
\newblock Supervised-learning-based development of multibit rcs-reduced coding
  metasurfaces.
\newblock \emph{IEEE Transactions on Microwave Theory and Techniques},
  70:\penalty0 264--274, 2022.

\bibitem[Goodfellow et~al.(2014)Goodfellow, Pouget-Abadie, Mirza, Xu,
  Warde-Farley, Ozair, Courville, and Bengio]{Goodfellow2014GenerativeAN}
Ian~J. Goodfellow, Jean Pouget-Abadie, Mehdi Mirza, Bing Xu, David
  Warde-Farley, Sherjil Ozair, Aaron~C. Courville, and Yoshua Bengio.
\newblock Generative adversarial nets.
\newblock In \emph{NIPS}, 2014.

\bibitem[CST(2021)]{CST}
{CST Studio Suite}, 2021.
\newblock URL \url{https://www.3ds.com/products-services/simulia/products/}.

\end{thebibliography}






\end{document}